\documentclass[10pt,journal,compsoc]{IEEEtran}
\usepackage{amsmath,amsfonts}

\usepackage{array}
\usepackage[caption=false,font=normalsize,labelfont=sf,textfont=sf]{subfig}
\usepackage{textcomp}
\usepackage{stfloats}
\usepackage{url}
\usepackage{verbatim}
\usepackage{graphicx}
\usepackage{cite}
\usepackage{multirow}
\usepackage{booktabs}

\usepackage{color}
\usepackage{hyperref}
\usepackage{graphicx}
\usepackage{amsmath}
\usepackage{amssymb}
\usepackage{cleveref}
\usepackage[section]{placeins}
\usepackage{float}
\usepackage{diagbox}

\usepackage{epstopdf}
\usepackage{amsmath}
\usepackage{amssymb}
\usepackage{multirow}
\usepackage{hyperref}

\usepackage{xcolor}

\usepackage{graphicx}%
\usepackage{multirow}%
\usepackage{amsmath,amssymb,amsfonts}%
\usepackage{amsthm}%
\usepackage{mathrsfs}%
\usepackage{xcolor}%
\usepackage{textcomp}%
\usepackage{manyfoot}%
\usepackage{booktabs}%
\usepackage{algorithm}%
\usepackage{algorithmicx}%
\usepackage{algpseudocode}%
\usepackage{listings}%

\usepackage{array}
\usepackage[caption=false,font=normalsize,labelfont=sf,textfont=sf]{subfig}
\usepackage{stfloats}
\usepackage{url}
\usepackage{verbatim}
\usepackage{color}
\usepackage{hyperref}
\usepackage{cleveref}
\usepackage[section]{placeins}
\usepackage{float}
\usepackage{diagbox}
\usepackage{epstopdf}
\usepackage [latin1]{inputenc}
\usepackage{bbding}
\usepackage{booktabs}

\ifCLASSINFOpdf
\else
\fi
\begin{document}
%
\title{An Open-World, Diverse, Cross-Spatial-Temporal Benchmark for Dynamic Wild Person Re-Identification}
%
%
%
%

\author{Lei~Zhang,
        Xiaowei~Fu, 
        Fuxiang~Huang,
        Yi Yang,
        Xinbo Gao
\IEEEcompsocitemizethanks{\IEEEcompsocthanksitem This work was partially supported by National Key R\&D Program of China (2021YFB3100800), National Natural Science Fund of China (62271090), Chongqing Natural Science Fund (cstc2021jcyj-jqX0023), and National Youth Talent Project. This work was also supported by Huawei computational power of Chongqing Artificial Intelligence Innovation Center. 

\IEEEcompsocthanksitem Lei Zhang, Xiaowei Fu and Fuxiang Huang are with the School of Microelectronics and Communication Engineering, Chongqing University, Chongqing 400044, China, and Peng Cheng Lab, Shenzhen, China.
(E-mail: leizhang@cqu.edu.cn, xwfu@cqu.edu.cn, huangfuxiang@cqu.edu.cn)

%

\IEEEcompsocthanksitem Yi Yang is with the College of Computer Science and Technology, Zhejiang University, Hangzhou 310058, China. (E-mail: yangyics@zju.edu.cn).

\IEEEcompsocthanksitem Xinbo Gao is with the Chongqing Key Laboratory of Image Cognition, Chongqing University of Posts and Telecommunications, Chongqing 400065, China. (E-mail: gaoxb@cqupt.edu.cn)}

\thanks{Manuscript received April 19, 2005; revised August 26, 2015.}}

%
%

\markboth{Journal of \LaTeX\ Class Files,~Vol.~14, No.~8, August~2015}%
{Shell \MakeLowercase{\textit{et al.}}: Bare Demo of IEEEtran.cls for Computer Society Journals}
%



\IEEEtitleabstractindextext{%
\begin{abstract}
Person re-identification (ReID) has made great strides thanks to the data-driven deep learning techniques. However, the existing benchmark datasets lack diversity, and models trained on these data cannot generalize well to dynamic wild scenarios. To meet the goal of improving the explicit generalization of ReID models, we develop a new \emph{O}pen-\emph{W}orld, \emph{D}iverse, Cross-Spatial-Temporal dataset named OWD with several distinct features. 1) \emph{Diverse collection scenes}: multiple independent open-world and highly dynamic collecting scenes, including streets, intersections, shopping malls, etc. 2) \emph{Diverse lighting variations}: long time spans from daytime to nighttime with abundant illumination changes. 3) \emph{Diverse person status}: multiple camera networks in all seasons with normal/adverse weather conditions and diverse pedestrian appearances (e.g., clothes, personal belongings, poses, etc.). 4) \emph{Protected privacy}: invisible faces for privacy critical applications. To improve the implicit generalization of ReID, we further propose a Latent Domain Expansion (LDE) method to develop the potential of source data, which decouples discriminative identity-relevant and trustworthy domain-relevant features and implicitly enforces domain-randomized identity feature space expansion with richer domain diversity to facilitate domain-invariant representations. Our comprehensive evaluations with most benchmark datasets in the community are crucial for progress, although this work is far from the grand goal toward open-world and dynamic wild applications.
\end{abstract}

}

\maketitle

\IEEEdisplaynontitleabstractindextext

%
\IEEEpeerreviewmaketitle

\begin{figure}[t]
\begin{center}
\includegraphics[width=0.9\linewidth,height=0.6\linewidth]{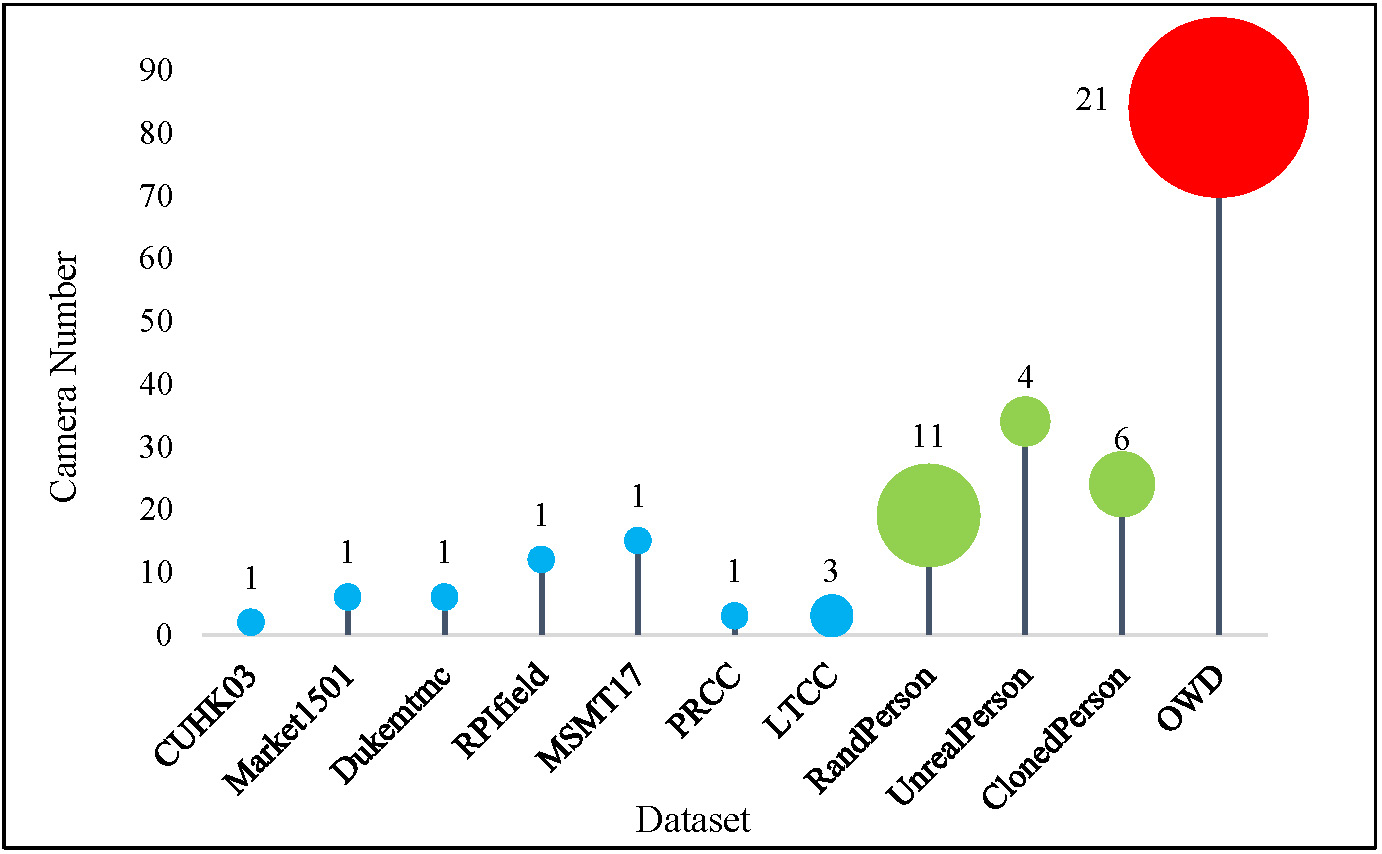}
\end{center}
\caption{Statistics of existing person ReID datasets. 
 The digit above each circle is the number of data collection scenes, the sky-blue circles are real data and the grass-green circles represent synthetic data}
\label{fig:image}
\end{figure}

\IEEEraisesectionheading{\section{Introduction}\label{sec1}}
\IEEEPARstart{D}{eep} Neural Network (DNN) has made undoubtedly great progress on most computer vision tasks in recent years. Person re-identification (ReID) is a fundamental task that exploits DNN in distributed multi-camera surveillance, which aims to match people across camera networks~\cite{huang2022learning, rao2019learning,li2020scalable}. Recently, the performance on some person ReID benchmarks has skyrocketed by assuming the training and test samples are drawn from the independent and identical distribution (i.i.d.) \cite{2020Salience, yin2020fine, zhang2020relation, zhu2019intra, zhai2023population}. However, there is often a serious performance degradation when a well-trained ReID model encounters out-of-distribution (OOD) test data with different styles~\cite{8578114} in applications. Collecting specific data and training DNNs for individual monitoring occasions might be a feasible strategy, but expensive, time-consuming and labor-intensive. Therefore, developing generalizable models that can be adapted to multiple scenarios without additional adjustment attract increasing attentions~\cite{jia2019frustratingly, 2019Generalizable, zhao2021learning}. However, the existing evaluation benchmarks are often collected from relatively closed-world scenes, which is insufficiently diverse towards the dynamic wild applications. Although some domain generalization strategies such as domain augmentation~\cite{shu2021open, zhang2020does} by changing domain style have been considered, the improvement is still limited due to the content diversity is overlooked. 
By empirically observing the success of computer vision tasks, diverse source data can always be a guarantee of developing powerful DNN models, which is also of great importance for generalizable person ReID in dynamic wild scenarios. 

\begin{figure*}[t]
\begin{center}
\includegraphics[width=0.9\linewidth,height=0.4\linewidth]{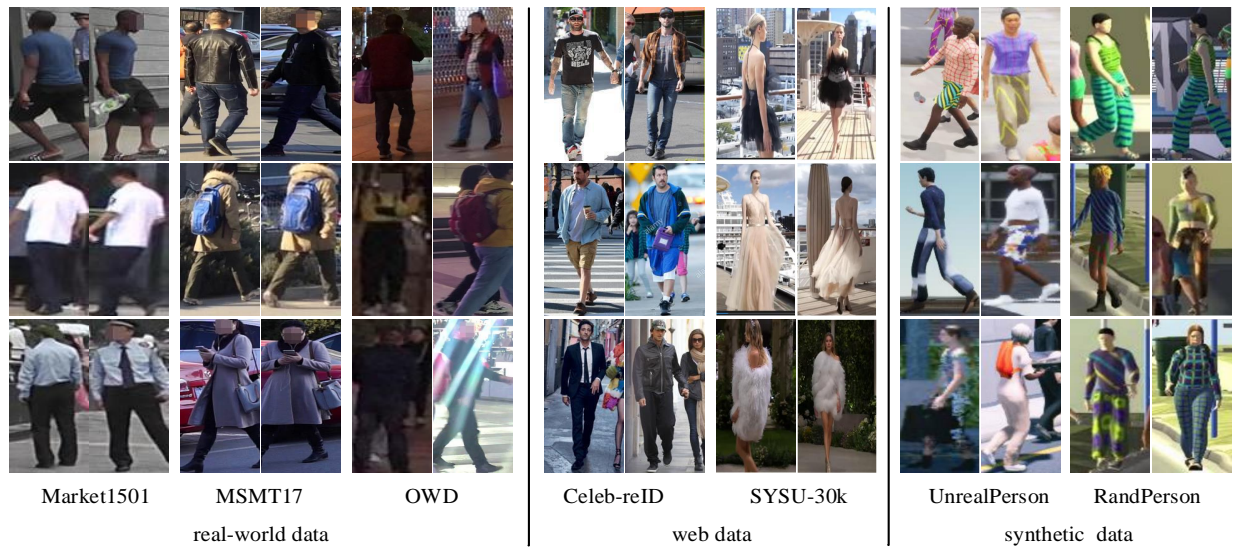}
\caption{Examples of the existing datasets from different sources (real-world data, web data and synthetic data). Compared to real-world benchmarks, OWD is more diverse and challenging. Web data is well-lit, clearer and finer, which limits its diversity. Large domain shift is observed between synthetic and real-world data. We mosaic the pedestrian faces obtained from real-world scenes without privacy issues
}
\label{samplecom}
\end{center}
\end{figure*}

The current pedestrian datasets mainly come from three sources: real-world collection \cite{zheng2015scalable, 8578114}, capturing from web videos or movies \cite{huang2019beyond, wang2020weakly} and synthesizing virtual characters \cite{zhang2021unrealperson, wang2022cloning}. Although rapid progress has been made in person ReID, previous algorithms mainly focus on these benchmarks with favorable conditions. As shown in Fig.~\ref{fig:image}, most publicly available person ReID datasets have following limitations. 1) The samples are collected from restricted scenes such as campus, which provide insufficient backgrounds and occlusions. 2) The original data only covers a short daytime range, which limits the diversity of low-illumination. 3) The scale of camera networks is small and merely deployed in one season, which restricts the variations of person appearance and weather conditions. 4) The privacy is not well protected due to the visible faces. These shortages limit the deployment of ReID in open-world scenarios and privacy-critical applications. For web benchmarks, large-scale data can be easily obtained, but lack diversity and challenges due to the demand of broadcast effect. As shown in Fig.~\ref{samplecom}, web images from film or videos are usually clearer and finer compared with real-world data. For synthetic benchmarks, although it is possible to generate infinite number of samples as needed, there exists large domain shifts across synthetic and real domains, which is unfavourable to domain generalization.

Based on the above observations, we present a new \emph{O}pen-\emph{W}orld, \emph{D}iverse, Cross-Spatial-Temporal dataset named OWD to meet the goal of person ReID in dynamic wild scenes. Compared with previous real-world benchmarks, OWD has the following significant strengths. 1) As shown in Fig.~\ref{fig:image}, samples are collected from more open-world scenes, such as shopping malls, streets, intersections, parks, etc. 
Multiple acquisition scenes naturally improve image content and domain diversity. 2) As shown in Fig.~\ref{samplecom}, the raw data span from daytime to nighttime, which significantly enriches the illumination variations. It is worth noting that although some existing benchmarks claim that they contain nighttime samples, they are either infrared data for cross-modality tasks \cite{2019Night} or from pictures with sufficient light in the film having little difference from daytime samples \cite{shu2021large}. 3) Multiple camera systems are deployed in all seasons, which supply diverse clothes, personal belongings, poses, etc. More cameras in open-world scenes ensure the diversity of the viewpoint of the samples and provide more rich backgrounds and content changes. 
Also, different from one-season benchmarks that are under normal weather conditions (favorable lighting and resolution), OWD covers all seasons in a year which helps improve generalization to adverse weather conditions and diverse pedestrian clothing, etc. 4) The visible faces are masked to meet privacy-critical applications. 5) Different from some existing datasets with labeling noises, OWD provides rich but high-quality annotations, including pedestrian ID, camera ID, scene ID, temporal information and day/night, which allows us to flexibly split samples for different applications. Furthermore, we set three evaluation protocols which allow the model to evaluate its generalization ability comprehensively under different domain gaps. Fig.~\ref{samplecom} shows the challenge of OWD, which is more in line with the open-world scenarios.

Since OWD contains abundant information, it is more conducive to train better models, but a fact is it is impractical to cover all inexhaustible open-world scenes. 
Domain generalization (DG) retrospected to a decade ago \cite{Blanchard2011GeneralizingFS} enforces domain-invariant feature learning on single/multiple source domains without access to target domain. Domain augmentation-based DG methods~\cite{shu2021open, zhang2020does} have been proven to improve model generalization by explicitly enriching the training data with different domain styles (i.e., batch statistics). 
Consider the relatively sufficient diversity of OWD and inspired by the explicit domain augmentation,
to further explore the potential of the data, we propose a simple but effective Latent Domain Expansion (LDE) method, which is substantially an implicit feature augmentation model. An intuitive idea behind LDE is to implicitly expand the breadth of source data from the perspective of enhancing domain diversity of identity-discriminative features without increasing training burden. Specifically, to improve the domain breadth and simultaneously guarantee the identity invariance, we propose to expand the latent feature space using Gaussian random variables characterized by domain-wise co-variances. The proposed LDE is composed of two steps: domain decoupling and domain expansion.

\textbf{Step 1: domain decoupling}. To separate domain-relevant and identity-relevant person features, a dual-stream network with multiple Domain Decouple Modules (DDM) is exploited. 
However, the decoupled domain-relevant features always imply some identity-relevant feature, which leads to inaccurate estimation of domain style and deteriorates the subsequent domain expansion. Therefore, a Mutual Similarity Lifting-Suppression (MSLS) strategy is proposed to purify the domain-relevant features from multiple levels of DDM by suppressing the identity-relevant parts implied in domain-relevant features and lifting other parts. 


\textbf{Step 2: domain expansion}. With the domain-wise co-variances computed via domain-relevant features, the identity feature space could be spanned by interactively adding a random direction sampled from a zero-mean Gaussian distribution characterized by the above domain-wise co-variance matrices. Furthermore, to reduce training burden, an $L_{\infty}$-cross-entropy loss with second-order Taylor expansion and the law of large numbers is derived.

The main contributions are summarized as follows.

\begin{itemize}
  \item We present an open-world person ReID benchmark, OWD, towards dynamic wild scenarios. OWD is collected from numerous scenarios with diverse changes in backgrounds, buildings, lighting conditions, weather conditions and clothing appearances, which provides a powerful benchmark for developing and evaluating the generalization potential of continually evolved models.
  \item To further improve open-world domain generalization, we propose a simple yet effective latent domain expansion (LDE) model, which implicitly expands the latent feature space with rich domain diversity of person identity based on domain-wise statistics and facilitates the domain-invariant representations.
  \item Numerous experiments and comparisons on a number of ReID datasets in the community verify that the transfer ability of our OWD significantly outperforms existing benchmarks. Also, the proposed LDE is comparable to state-of-the-art domain generalization (DG) ReID models.
\end{itemize}

\section{Related Work}
\subsection{Datasets of Person ReID}

\textbf{1) Real-World ReID datasets.} With the development of deep learning, many person ReID datasets have been collected from the real world, including VIPeR \cite{2008Viewpoint}, GRID \cite{2014Person}, CUHK01-03 \cite{6619305,2012Human,6909421}, Market1501 \cite{zheng2015scalable}, Dukemtmc \cite{zheng2017unlabeled} and MSMT17 \cite{8578114}. A large-scale Person30K~\cite{2021Person30K} dataset is collected to improve model generalization. However, there is no nighttime data which is quite important in practical application. These previous datasets have some limitations on collection scenes, camera networks and lighting conditions, and 
limit the generalization potential of the developed model. In addition, Fu et al. propose LUPerson \cite{fu2020unsupervised} to perform unsupervised pre-training for person ReID. Although it is large in scale, it does not have pedestrian label.

\textbf{2) Web ReID datasets.} Some recent work collects pedestrian data from videos and movies on the web. PIPA \cite{zhang2015beyond} is collected from public photo albums and consists of 2,356 different pedestrians. CIM \cite{huang2018unifying} is collected from movies for person retrieval and person search, which contains 1,218 identities but not cropped to a single pedestrian. Images of Celeb-reID \cite{huang2019beyond} are captured from the Internet using street snap-shots of celebrities, which contains a total of 1,052 identities with 34,186 images in Celeb-reID. SYSU-30k \cite{wang2020weakly} is a large-scale ReID benchmark collected from web and TV programs. However, its annotation is bag-level and not accurate enough. LaST \cite{shu2021large} is collected from videos and movies and presents more challenging and high-diversity ReID settings, but is mainly used for cloth-changing ReID task. Although it is convenient to collect data from the web, pedestrians in movies and videos are often in a well-lit environment, which limits the lighting diversity of data. In addition, for artistic effects, tools such as filters may cause domain gaps between these data and real scenes.

\textbf{3) Synthetic ReID datasets.} Some recent work focuses on the generation of synthetic data due to the difficulty of collecting and labeling real-world data. SOMAset \cite{Barbosa2018Looking} is created by photorealistic human body generation software. The training data consists of a synthetic 100K instances. PersonX \cite{sun2019dissecting} is a large-scale controllable synthetic data engine that could synthesize pedestrians by setting the visual variables to arbitrary value. RandPerson \cite{wang2020surpassing}  simulates a number of different virtual environments using Unity3D and develops an automatic code to randomly generate various different 3D characters. More recently, UnrealPerson \cite{zhang2021unrealperson} generates synthesized images of high quality from controllable distributions. Based on UV texture mapping, ClonedPerson \cite{wang2022cloning} designs two cloning methods to clone the whole outfits from real-world person images to virtual 3D characters. Although these datasets can possess almost countless identities and samples, there exists a large domain discrepancy between synthetic and real-world data.


\begin{figure*}[t]
\begin{center}
\includegraphics[width=0.9\linewidth,height=0.8\linewidth]{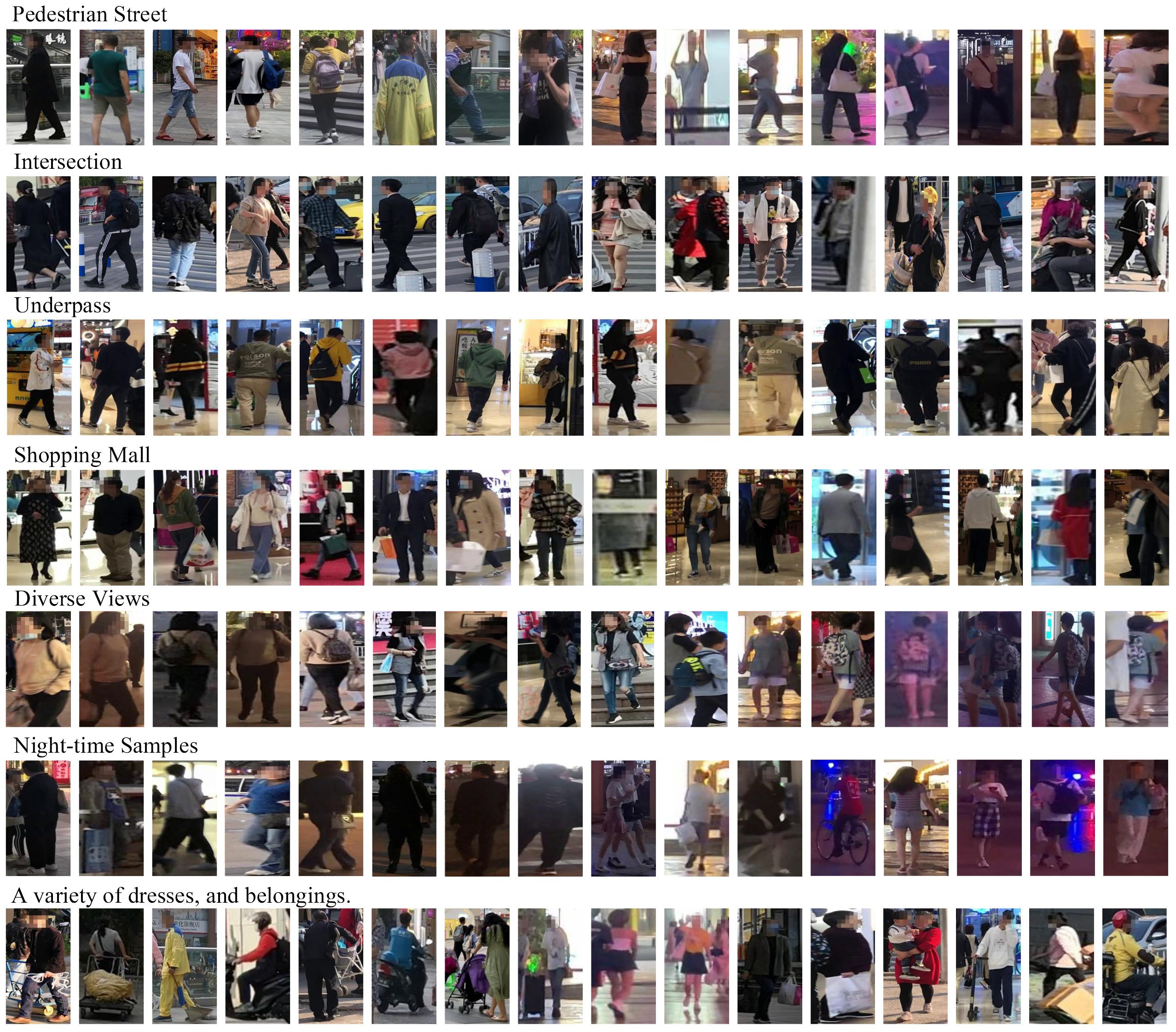}
\caption{Examples from OWD with various scenes, camera views, lighting conditions and clothing changes, etc 
}
\label{datafeat}
\end{center}
\end{figure*}

\subsection{Domain Generalization for Person ReID}

Domain Generalization (DG) aims to learn a domain-invariant model based on single or multiple source domains without access to target domain, which attracts attention from person ReID community. 
DualNorm \cite{jia2019frustratingly} employs both Instance Normalization (IN) and Batch Normalization (BN) to eliminate domain style. \cite{9157711} propose a style normalization and restitution module to distill identity-relevant features to bypass domain discrepancy. DDAN \cite{chen2021dual} tries to seek central and peripheral domains with minimal distribution shifts. \cite{zhao2022revisiting} propose stochastic splitting-sliding sampler and variance-varying gradient dropout to force model to jump out of local minimum under stochastic sources. \cite{9157711} and \cite{qian2020long} follow a similar domain decoupling pipeline: normalization and then restitute from residual features. However, these domain decoupling methods neglect the utilization of domain information, and the proposed LDE focuses on obtaining purer domain features for more effective domain expansion.

Meta-learning-based strategy is also popular in DG ReID. \cite{2021Person30K} design a learning then generalization evaluation meta-training procedure and a meta-discrimination loss. 
DIMN \cite{2019Generalizable} learns the classifier from gallery images and computes matching scores between gallery and the probe images by following the idea of meta-learning. M$^{3}$L \cite{zhao2021learning} simulates the train-test process of domain generalization for learning domain-invariant models. MetaBIN \cite{choi2021meta} aims at generalizing traditional normalization layers by simulating unsuccessful generalization scenarios beforehand in the meta-learning pipeline. However, meta-learning based DG approaches usually increase the optimization complexity.

Besides, domain augmentation is widely-used in DG image classification tasks. \cite{zhang2020does} prove that Mixup augmentation is a type of data-adaptive regularization to reduce overfitting. \cite{li2021simple} find that perturbing
the feature embedding with Gaussian noise during training brings domain generalization capability to a classifier. \cite{shu2021open} augment domains on feature-level by a new Dirichlet mixup and label-level by distilled soft-labeling. For DG person ReID, it may be time-consuming to generate large-scale fine pedestrian images. ISDA~\cite{wang2019implicit} achieves intra-class feature-level implicit data augmentation for the classification task. Different from the above algorithms, inspired by ISDA, we propose Domain Expansion in LDE to conduct latent domain-wise expansion to domain-invariant feature learning for the DG ReID task. 

\section{The OWD Dataset}

\subsection{Description}

\begin{table*}[t]
\caption{Comparison between some publicly available ReID datasets and our OWD}
\renewcommand{\arraystretch}{0.5}
\setlength{\tabcolsep}{0.6mm}
\begin{tabular*}{\textwidth}{@{\extracolsep\fill}lcccccccc}
\toprule
Datasets & scene & sites & cameras & night data  & seasons  & identity privacy & images & identities  \\
\midrule
VIPeR & - & 1 & 2 & \XSolidBrush  & 1 & \XSolidBrush & 1,264 & 632\\
iLIDS & airport & 1 & 2 & \XSolidBrush  & 1 & \XSolidBrush & 476 & 119\\
GRID & subway & 1  & 8 & \XSolidBrush  & 1 & \XSolidBrush &1,275 & 1,025\\
CAVIAR4 & market & 1 & 2 & \XSolidBrush  & 1 & \XSolidBrush & 1,220 & 72\\
3DPeS & outdoor & 1 & 8 & \XSolidBrush  & 1 & \XSolidBrush & 1,011 & 192\\
PRID & road & 1 & 2 & \XSolidBrush  & 1  & \XSolidBrush & 1,134 & 934\\
WARD & - & 1  & 3 & \XSolidBrush  & 1  & \XSolidBrush & 4,786 & 70\\
CUHK03 & campus & 1  & 6 & \XSolidBrush  & 1 & \XSolidBrush &14,097 & 1,467\\
Market1501  & campus& 1  & 6 & \XSolidBrush  & 1   & \XSolidBrush & 32,668 & 1,501\\
PKU-Reid  & - & 1  & 2 & \XSolidBrush  & 1   & \XSolidBrush & 1,824 & 114\\
PRW  & campus& 1  & 6 & \XSolidBrush  & 1   & \XSolidBrush & 32,668 & 1,501\\
Dukemtmc&  campus & 1  & 6 & \XSolidBrush  & 1   & \XSolidBrush & 34,304 & 932\\
MSMT17  & campus& 1  & 15 & \XSolidBrush & 1 & \XSolidBrush & 126,441 & \textbf{4,101}\\
RPIfield &  outdoor& 1  & 12 & \XSolidBrush  & 1   & \XSolidBrush & \textbf{601,581} & 112\\
PIPA &  web& -  & - & -  & -   & \XSolidBrush & 37,107 & 2,356\\
CIM &  web& -  & - & -  & -   & \XSolidBrush & 72,875 & 1,218\\
Celeb-reID &  web& -  & - & -  & -   & \XSolidBrush & 34,186 & 1,052\\
Celebrities-reID  & web& -  & - & - & - & \XSolidBrush & 10,842 & 590\\
PRCC  & teaching room & 1  & 3 & - & - & \XSolidBrush & 33,698 & 221\\
LTCC & office building & 3  & 3 & - & - & \XSolidBrush & 17,138 & 152\\
\textbf{OWD} & \textbf{open} & \textbf{21}  & \textbf{84} & \Checkmark  & \textbf{4}  & \Checkmark & 136,614 & 3,986\\
\hline
\end{tabular*}
\label{data compa}
\end{table*}

\begin{figure}[t]
\centering
\captionsetup[subfloat]{labelfont=scriptsize,textfont=scriptsize}
\subfloat[The number of identities and samples on each scene.]{
\begin{minipage}[t]{1.0\linewidth}
\centering
\includegraphics[width=1.0\linewidth]{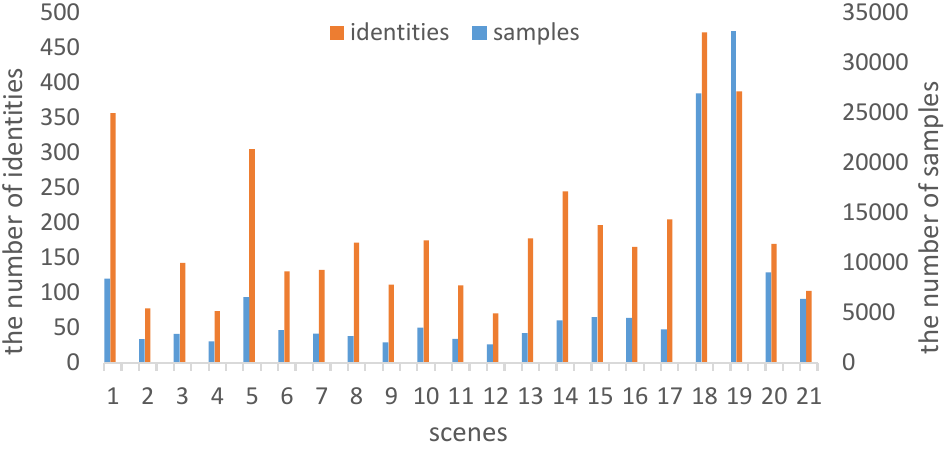}
\end{minipage}%
}%

\subfloat[Images of day and night.]{
\begin{minipage}[t]{0.5\linewidth}
\centering
\includegraphics[width=0.8\linewidth]{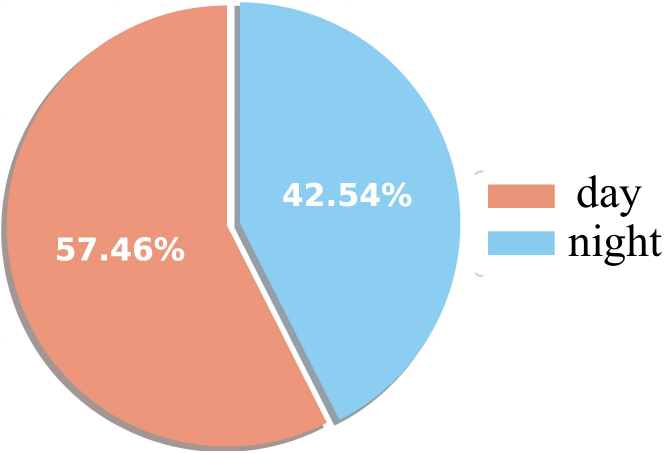}
\end{minipage}
}%
\subfloat[IDs of day and night.]{
\begin{minipage}[t]{0.5\linewidth}
\centering
\includegraphics[width=0.8\linewidth]{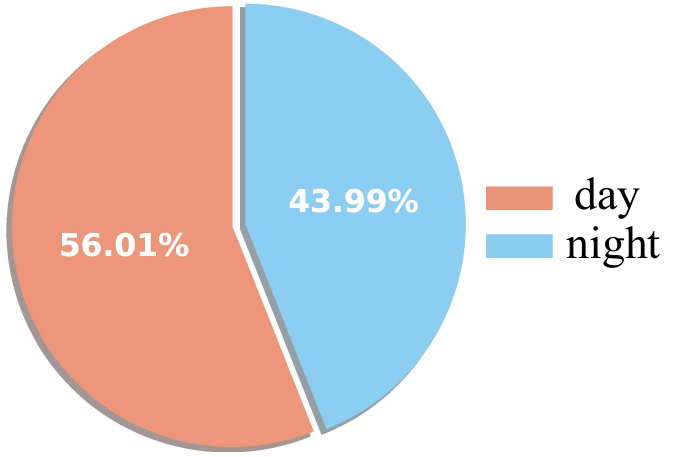}
\end{minipage}
}%

\centering
\caption{Statistics of the collected OWD}
\label{fig:sta}
\end{figure}

To promote the practical application of ReID models in dynamic wild scenarios, we consider the diversity of scenes, illuminations and seasons to construct a challenging dataset. First, we select multiple disjoint scenes (21 scenes) involving both indoor and outdoor sites with various backgrounds and occlusions, etc. Second, we make the first attempt to introduce real-world nighttime samples for improving model adaptability to adverse low-lighting conditions. Third, a large-scale camera system (84 cameras) is deployed in all seasons to capture challenging and practical person samples under adverse weather conditions. Besides, the visible face is masked to deal with privacy-critical issues. We annotate the data manually to eliminate the label noise from a pre-trained pedestrian detector. During the annotation process, each annotator is responsible for the data of a location. After annotation, different annotators exchange label results to ensure the quality of labels. To enhance the challenge and diversity of OWD, we sample images from the captured videos at a longer time stamp interval and manually filter those samples that are adjacent in time stamp and visually similar during labeling. The proposed OWD dataset totally contains 136,614 bounding boxes of 3,986 identities. Although OWD has no the largest size in data scale compared with some large-scale benchmarks, eliminating ``redundant" samples effectively makes OWD diverse and increases its challenge. Each pedestrian in OWD crosses at least two scenes in one collection region to ensure intra-class diversity. Compared with the publicly available datasets, OWD aims to introduce more challenging factors considering the practical application scenarios, as shown in Fig.~\ref{datafeat}, rather than very large-scale data with redundance. Data statistics for comparing with more datasets is shown in Table~\ref{data compa}, such as VIPeR \cite{2008Viewpoint}, iLIDS \cite{2009Associating}, GRID \cite{2014Person}, CAVIAR4 \cite{cheng2011custom}, 3DPeS \cite{baltieri20113dpes}, PRID \cite{Hirzer2011PersonRB}, WARD \cite{martinel2012re}, CUHK03 \cite{6909421}, Market1501 \cite{zheng2015scalable}, PKU-Reid \cite{ma2016orientation}, PRW \cite{zheng2017person}, Dukemtmc \cite{zheng2017unlabeled}, MSMT17 \cite{8578114}, RPIfield \cite{zheng2018rpifield}, PIPA \cite{zhang2015beyond}, CIM \cite{huang2018unifying}, Celeb-reID \cite{huang2019beyond}, Celebrities-reID \cite{huang2019celebrities}, PRCC \cite{yang2019person}, and LTCC \cite{qian2020long}. We observe that the OWD provides personalized new features such as nighttime data, large-scale camera networks and open scenes, etc. to ensure temporal-spatial diversity. Detailed statistic about our OWD is illustrated in Fig.~\ref{fig:sta}. We summarize the distinctive traits of the OWD as follows.

\begin{figure*}[t]
\begin{center}
\includegraphics[width=0.9\linewidth,height=0.7\linewidth]{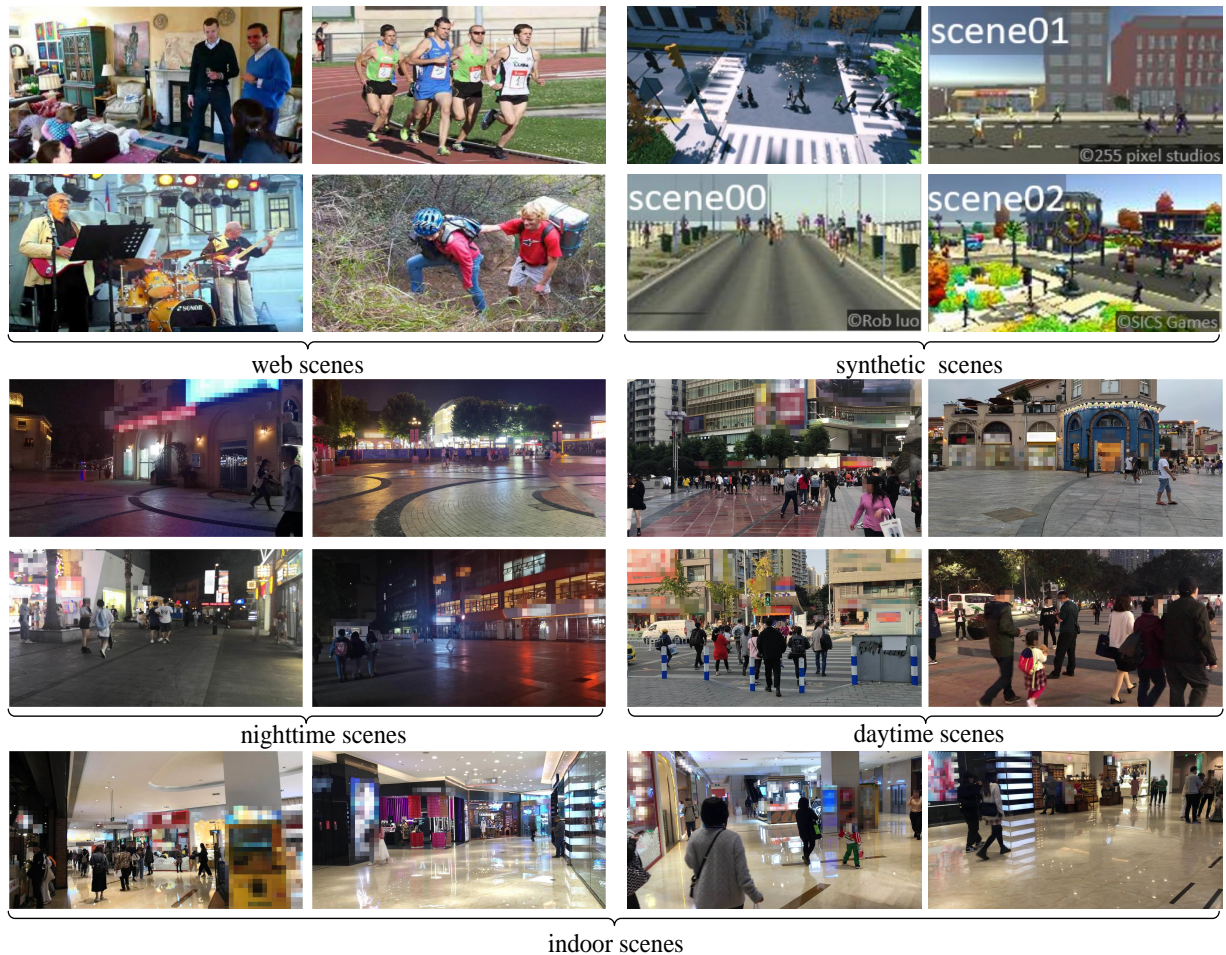}
\end{center}
   \caption{Several collection scenarios of web data, synthetic data and our OWD. Web scenes are usually clear and refined owing to their broadcasting purpose. Moreover, due to a large number of close-up shots, pedestrian samples tend to have high resolution. There is a large domain gap between synthetic data and the real-world data. OWD is more diverse. To avoid privacy issue, mosaic is provided}
\label{sta}
\end{figure*}

\textbf{1) The collection scenes are open-world.} We capture pedestrian images from 21 sites that do not intersect each other, such as shopping malls, pedestrian streets, underpasses, intersections, squares, etc., involving both indoor and outdoor sites. As shown in Fig.~\ref{datafeat}, multiple scenes in open-world environment make the appearance of pedestrian samples diverse and the background and lighting changes are more abundant, because 21 sites are alternatedly experienced across different time stamps in different seasons. Previous real-world benchmarks are usually only collected in very few location, such as campus or classroom. Although they provide large-scale samples, the limited collection scenarios restrict the diversity and not conducive to dynamic wild person re-id.
Fig.~\ref{sta} shows the comparison of collection scenarios in different types of data.
For the types of web and synthetic datasets, although various collection scenarios can be easily accessed, they have serious shortcomings compared to real-world data, elaborated as follows.

For web data, with the development of camera sensors, to give audiences a better visual experience, videos or movies on the web are well-lit and have high legibility. Moreover, due to close-up shooting, the resolution of these pedestrian images is usually high. The above factors mean low diversity of web data, and naturally the data challenge is reduced. That is, web data cannot provide corresponding tough scenarios for developing generalizable models adapted to adverse scenes, where a real-world monitoring system often meets. 

For synthetic data, there are non-negligible domain differences between synthetic and real data. Besides, the illumination distribution of the synthetic scene is uniform and monotonous, which limits the lighting diversity. The proposed OWD contains diverse, cross temporal-spatial scenes, from daytime to nighttime and from outdoor to indoor, etc. Due to the impact of background, lighting and other factors, the appearance of pedestrians in different scenes is significantly different. This greatly improves the diversity of OWD. Unlike web data that focuses on the specific protagonists, OWD is collected from dynamic wild scenes and can capture pedestrians of different statuses. This significantly improves the intra-class diversity of pedestrians, making the data benchmark more challenging.

\textbf{2) Nighttime person samples are collected.} We provide person ReID data from RGB sensors at nighttime for the first time. On the one hand, compared with daytime scenes, pedestrian images lose many details due to insufficient lighting at night, which increases the matching difficulty between pedestrians. From Fig.~\ref{datafeat}, we observe that the person appearance at nighttime collected from different sites varies quite drastically, which provides challenging situations. On the other hand, even for the same pedestrian, the appearance from various angles is quite different in low lighting condition. For example, when a pedestrian in red clothing is captured from the angle of the side light or backlight, the color of clothes may look closer to black. This increases intra-class variances significantly and conducive to learning identity-invariant features. Fig.~\ref{datafeat} and Fig.~\ref{sta} show the nighttime examples in different collection scenarios. It is worth noting that although some benchmarks claim that they collect nighttime samples, they are either infrared data for cross-modality tasks \cite{2019Night} or data with sufficient illumination having little disparity from daytime data \cite{shu2021large}. 

\begin{figure}[t]
\begin{center}
\includegraphics[width=0.9\linewidth,height=0.5\linewidth]{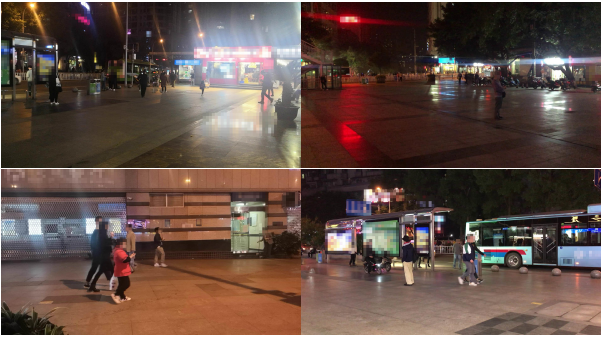}
\end{center}
   \caption{Examples of pictures captured by different cameras under the same collection scene, which show rich visual changes}
\label{camera_scene}
\end{figure}

\textbf{3) The seasons are diversified.} A majority of previous real-world benchmarks are short-term data, such as Market1501 \cite{zheng2015scalable}, MSMT17 \cite{8578114}, etc. Their collection time is concentrated, which limits the temporal diversity. The proposed OWD is acquired from all seasons continuing two years, so the clothes and personal belongings that are of utmost importance to ReID can be in various styles. Also, compared with previous single-season benchmarks, the all-seasons setting of OWD could supply various adverse conditions including poor weather (e.g., foggy, rainy, etc.) and lighting to strengthen the ReID models. As shown in Fig.~\ref{datafeat}, pedestrian samples under adverse conditions clearly raise the challenge of detection and matching.

\textbf{4) The scale of the camera system is expanded.} OWD collects data from a total of 84 cameras to capture abundant variations from different viewpoints. The proposed OWD covers a variety of pedestrian poses besides walking, such as riding a motorcycle or pushing a baby carriage. Also, OWD contains diverse personal belongings, such as briefcases, shopping bags and tools. Under each open-location, four cameras are deployed to obtain images of the same pedestrian from multiple views. As shown in Fig.~\ref{camera_scene}, even in the same scene, these four cameras can provide quite different scene characteristics, including background, lighting and occlusion. This significantly expands the intra-class differences within the same identity.

\textbf{5) The annotations are abundant.} To facilitate model training in the future, we provide abundant annotations including pedestrian ID, camera ID, scene ID, temporal information, and day/night annotations. These annotations play a significant role in future research on generalizable person ReID models in dynamic wild scene.

\textbf{6) The visible faces are masked.} The face privacy information is masked to deal with privacy-sensitive issues. Most previous benchmarks have visible faces. Also, with identity-related faces masked, it brings more difficulty for discriminative identity feature learning.

\textbf{7) Multi-level evaluation protocols.} Benefiting from rich annotations, the training/test division of the proposed OWD can be flexibly managed. OWD provides multiple splits at three levels based on domain gap, for comprehensive evaluation of a model. These protocols are elaborated as below.

\begin{figure}[t]
\begin{center}
\includegraphics[width=0.9\linewidth,height=0.5\linewidth]{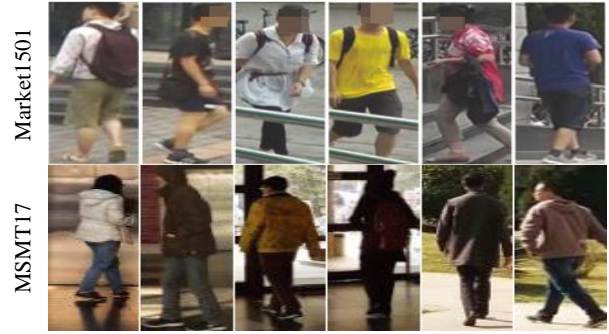}
\end{center}
   \caption{The \textbf{close-scene} setting of previous datasets. 
   The odd columns mean training data and the even columns mean test data. The training and test data have the same background.}
\label{otherdata}
\end{figure}

\begin{table}[t]
\centering
\caption{The training/test splits of OWD under different evaluation protocols}\label{strategy}
\setlength{\tabcolsep}{1.2mm}
\begin{tabular}{lcccccc}
\toprule%
& \multicolumn{2}{c@{}}{Identities} & \multicolumn{2}{c@{}}{Samples} & \multicolumn{2}{c@{}}{Cameras}\\\cmidrule{2-3}\cmidrule{4-5}\cmidrule{6-7}%
Protocol & Train & Test & Train & Test & Train & Test  \\
\midrule
close-scene & 951 & 1,902 & 20,700 & 40,412 & 68 & 68 \\
open-scene & 1,474 & 2,512 & 64,358 & 72,589 & 32 & 52 \\
day-night & 1,598 & 1,283 & 35,116 & 37,158 & 40 & 32 \\
\hline
\end{tabular}
\end{table}

\subsection{Evaluation Protocols}
\label{sect_eval}
In previous datasets, we notice that different people in the training set and the test set usually appear under the same scene as shown in Fig. \ref{otherdata}, which simplifies the ReID task because the backgrounds or occlusions may have been fitted during training due to dataset bias. We call this setup \textbf{close-scene}. However, in practical application, it is unrealistic to collect samples for training each monitoring system separately. Benefiting from richer annotations (such as scenes, cameras, temporal information, daytime/nighttime, etc.), OWD could be managed flexibly with two extra new evaluation protocols: \textbf{open-scene} and \textbf{day-night} protocols, rather than only close-scene to facilitate the comprehensive evaluation of ReID models, 
as shown in Table~\ref{strategy}. These two new protocols are related to larger domain discrepancies and are therefore more challenging. 

\textbf{1) Open-scene} protocol. Under this setting, the training and test data are from disjoint sites. Besides, the person identity, background, occlusion, and lighting of training and test data do not overlap. Compared with close-scene setting, this open-scene protocol is more challenging and realistic. Under the open-scene protocol, the model needs to overcome the test-time interference factors (lighting, background, etc.) that have not been previously accessed in training to capture the discriminative pedestrian features.

\textbf{2) Day-night} protocol. The training set is composed of the day-time images, while the test samples are sampled from nighttime scenarios in different sites. This setting provides a benchmark to verify the generalization ability of models when encountering serious lighting domain gap. Actually, in outdoor sites, there is a huge appearance difference of pedestrian data between day and night, not to mention other interference of background and occlusion under quite different lighting conditions.

Overall, OWD provides three evaluation protocols: \textbf{close-scene}, \textbf{open-scene} and \textbf{day-night}. The identity and sample division of the training/test set are shown in Table~\ref{strategy}. Under these three different settings, the performance of a model can be comprehensively evaluated at different domain gaps. In OWD, each query image has multiple positive samples in the gallery set. Therefore, similar to most previous datasets, the mean average precision (mAP) is used to evaluate the overall performance. 

%

\subsection{Open Problems in OWD}

As a new diverse benchmark, OWD could provide new exploration space for several fundamental computer vision tasks such as pedestrian detection, tracking and re-identification in open-world scenes. In this section, we discuss several open issues that OWD may bring to ReID community. First, since the diversity of OWD is sufficient, it is more conducive to single-source domain generalization of ReID without leveraging other auxiliary source domains. 
Second, since OWD supports editable person annotations, weakly-supervised person ReID with label noises can be easily studied. Third, OWD can be used for effective pre-training, which is beneficial to some scenarios where data collection is difficult, such as unsupervised person ReID. Finally, OWD also provides time stamps for person samples, and thus the temporal information can be exploited as prior knowledge for improving ReID performance. 

\subsection{The Difference between OWD and Combined Datasets}

As a new diverse benchmark, OWD has significant differences compared with the combination of multiple existing datasets. On the one hand, OWD provides new diversity, such as the setting up of night scenes, diverse seasons, etc., which are key for DG but cannot be obtained even when combining multiple public datasets. On the other hand, OWD captures data from wild and realistic scenes by multiple cameras, providing abundant intra-class variations and assisting the model in learning discriminative features. However, previous data was usually collected from simple and limited environments such as classrooms, this resulted in limited intra-class variations even if combining these data. Further, combining OWD and other datasets may demonstrate greater generalization potential for DG ReID task.

\section{The Proposed LDE Model}

\begin{figure*}[t]
\begin{center}
\includegraphics[width=0.9\linewidth,height=0.5\linewidth]{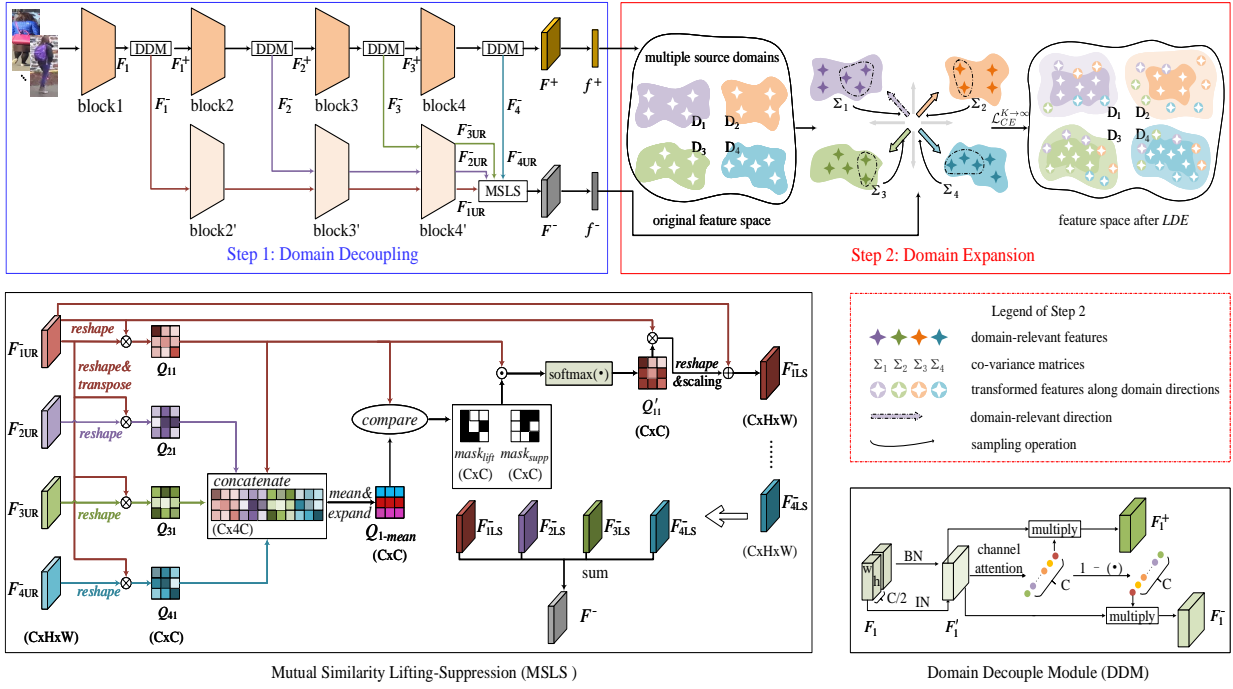}
\end{center}
   \caption{Illustration of LDE, which consists of two steps: Domain Decoupling and Domain Expansion. In Step 1, a dual-stream network with DDMs and MSLS is exploited to decouple identity-relevant ($f^+$) and domain-relevant features ($f^-$). In Step 2, the domain expansion is enforced to generate new identity features by adding a random variable sampled from a Gaussian distribution characterized by zero mean and domain-wise co-variance (i.e., domain style). Then, an implicit infinite cross-entropy loss is derived for training}
\label{DIFA}
\end{figure*}

\subsection{Problem Definition}
Suppose there are $S$ different source domains, $\left\{{D}_{1}, \ldots, {D}_{S}\right\}$, where ${D}_{j}=\small{\left\{\left(\boldsymbol{x}_{i}^{j}, y_{i}^{j}\right)\right\}_{i=1}^{N_{j}}}$, and 
${N}_{j}$ is the number of images in domain ${D}_{j}$. Each sample $\boldsymbol{x}_{i}^{j}$ is associated with its identity $y_{i}^{j} \in \mathcal{Y}_{j}=\left\{1,2, \ldots, M_{j}\right\}$, where ${M}_{j}$ is the number of person identities in domain ${D}_{j}$. DG ReID enables the model trained on multiple source domains to adapt to open target domains.

\subsection{Latent Domain Expansion}
\subsubsection{Overview}
Substantially, OWD is an explicit expansion in image level with sufficient diversity. Consider the fact that open-world data is always inexhaustible, to promote the generalization of ReID models, we further propose a \emph{L}atent \emph{D}omain \emph{E}xpansion (LDE), from the perspective of implicit domain expansion in feature level.
As shown in Fig.~\ref{DIFA}, LDE consists of two steps: Domain Decoupling and Domain Expansion. In Step 1, discriminative identity-relevant and trustworthy domain-relevant features are captured by a dual-stream model with several domain decouple modules (DDM) and a mutual similarity lifting-suppression (MSLS) module. Domain-relevant features are used to compute the domain-wise co-variance matrices used for subsequent domain expansion. In Step 2, domain expansion is achieved by adding random domain variables (potential semantic directions) to generate new features. The random variable is sampled from a zero-mean multi-variable normal distribution characterized by the computed domain-wise co-variances in step 1. To avoid extra training burden brought by randomized features, the above procedure is derived as an $L_{\infty}$-cross-entropy loss via second-order Taylor expansion and the law of large numbers. Through the above two steps, LDE is enforced to facilitate domain invariant  learning.
\subsubsection{Domain Decoupling}
We deploy a dual-stream network that does not share parameters and add multiple Domain Decoupling Modules (DDM) on different depths to initially separate identity-relevant and domain-relevant features. Then a mutual similarity lifting-suppression (MSLS) module is deployed to purify the domain-relevant features by removing some potential identity-relevant features. Take the widely used ReID network backbone, ResNet-50, as an example, DDM is added after each convolutional block.

\textbf{Details of DDM.} Take the DDM after block 1 as an example, we feed the feature map ${F}_{1}$ as input. Due to the impact of illumination and viewpoint, there is a large difference among images. To avoid the instability caused by drastic appearance changes of images during domain-relevant feature estimation, we exploit instance normalization ($\operatorname{IN}$) to alleviate the bias of the intense instance-relevant information.  $\operatorname{IN}$ can excessively eliminate domain information, so we further employ batch normalization ($\operatorname{BN}$) as a remedial measure. $\operatorname{BN}$ associates multiple features of the domains and realizes information interaction, which is beneficial to decoupling domain-relevant features. But in practical application, we empirically observe the performance is not sensitive to how to exploit the above two normalization methods. 
Inspired by \cite{pan2018two}, we conduct $\operatorname{IN}$ on the first half of the channels and perform $\operatorname{BN}$ on the last half to save memory. Then normalized feature ${F_{1}^{'}}$ could be obtained.

Given ${F_{1}^{'}}$, we further decouple it into two parts: domain-relevant feature ${F_{1}^{-}}$ and identity-relevant feature ${F_{1}^{+}}$. Specifically, we leverage the SE-like \cite{8578843} channel attention vector $\mathbf{\textbf{a}}=\left[a_{1}, a_{2}, \cdots, a_{d}\right] \in \mathbb{R}^{C}$ to mask ${F_{1}^{'}}$ . Then the decoupled features can be written as
\begin{equation} \label{augfeat}{F_{1}^{+}}= \mathbf{\textbf{a}} *  {F_{1}^{'}}, \;{F_{1}^{-}}= (\mathbf{1 - \textbf{a}})* {F_{1}^{'}}  \end{equation}

\begin{figure}[t]
\begin{center}
\includegraphics[width=0.9\linewidth,height=1.1\linewidth]{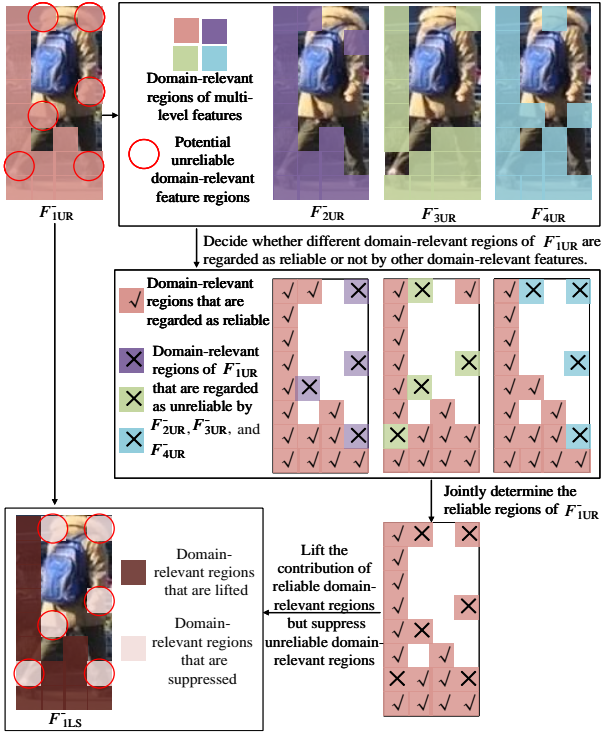}
\end{center}
   \caption{The main idea of MSLS. Take ${F_{1UR}^{-}}$ as an example, to suppress its ambiguous domain-relevant features (i.e., dominated by identity-relevant information), the agreement degree with the multi-level domain-relevant features ${F_{2UR}^{-}}$, ${F_{3UR}^{-}}$ and ${F_{4UR}^{-}}$ is computed. The domain-relevant features are substantially purified by lifting reliable pixels and suppressing unreliable pixels indicated by a mask
   }
\label{MSLS}
\end{figure}

Similarly, we can obtain $\{F_{i}^{+}, F_{i}^{-}\}$, where $i=2,3,4$ means the block index. However, DDM can only initially decouple domain-relevant and identity-relevant features, and the domain-relevant features from DDM still contain ambiguous domain information (i.e., dominated by identity-relevant information), as shown in the red circles in Fig.~\ref{MSLS}. Therefore, MSLS is introduced to purify the domain-relevant feature in multiple levels from DDMs, such that more accurate domain style can be computed.

\textbf{Details of MSLS.} As shown in Fig.~\ref{DIFA}, ${F_{1UR}^{-}}$, ${F_{2UR}^{-}}$ and ${F_{3UR}^{-}}$ are high-dimensional mappings of domain-relevant features ${F_{1}^{-}}$, ${F_{2}^{-}}$ and ${F_{3}^{-}}$, respectively, and ${F_{4UR}^{-}}={F_{4}^{-}}$, which are fed into the MSLS module for domain-relevant feature purification by removing the ambiguous domain information (i.e., identity-relevant). The basic idea of MSLS is described in Fig.~\ref{MSLS}.

Take ${F_{1UR}^{-}}$ as an example, to obtain purified domain-relevant information for domain expansion, 
its \textit{agreement} degree with the multi-level domain-relevant features $F_{2UR}^{-}$, $F_{3UR}^{-}$ and $F_{4UR}^{-}$ 
is used to determine whether each region of ${F_{1UR}^{-}}$ is domain trustworthy or ambiguous. Inspired by attention mechanism \cite{8578843, 2018CBAM}, its \textit{agreement} degree with multi-level domain-relevant features from DDM is measured by channel-wise mutual similarity. 
Only the regions having consistent agreement by multi-level domain-relevant features from DDMs are considered trustworthy (reliable), otherwise, ambiguous (unreliable). Then, the importance of those trustworthy regions is lifted while that of ambiguous regions is suppressed. 

The detailed flow of MSLS is shown in Fig.~\ref{DIFA}. For ${F_{1UR}^{-}}$, firstly, its channel-wise self-similarity is indicated as ${Q_{11}}$ and its mutual similarity between ${F_{1UR}^{-}}$ and other domain-relevant features ${F_{2UR}^{-}}$, ${F_{3UR}^{-}}$ and ${F_{4UR}^{-}}$ are calculated as ${Q_{21}}$, ${Q_{31}}$ and ${Q_{41}}$ $ \in \mathbb{R}^{C \times C}$, respectively. $C$ denotes the channel number. The above similarity matrices represent the agreement degree. 
The first row of ${Q_{21}}$ denotes the agreement degree with ${F_{2UR}^{-}}$ at the first channel of ${F_{1UR}^{-}}$. Then, these similarity matrices are concatenated and averaged in row-wise represented as $Q_{1-mean}$, used to capture the overall agreement degree for each channel of ${F_{1UR}^{-}}$. By conducting an element-wise comparison between ${Q_{11}}$ and $Q_{1-mean}$, the reliable pixels (i.e., trustworthy domain-relevant information) are obtained as a mask, \begin{equation} \label{masklift}{mask_{lift}[i,j]}=\begin{cases}
 1, \ if \ {Q_{11}[i,j]}\ge {Q_{1-mean}[i,j]}\\
0,\ if\ {Q_{11}[i,j]}<  {Q_{1-mean}[i,j]}
\end{cases} \vspace{-0.1cm} \end{equation}The ambiguous pixels are obtained as $mask_{supp}=1-mask_{lift}$.

To lift the contribution of reliable pixels while suppressing ambiguous pixels in ${F_{1UR}^{-}}$, the masked ${Q_{11}}$  by $mask_{lift}$ and $mask_{supp}$ are fed into the $softmax(\cdot)$ 
to obtain a new attention matrix $Q_{11}^{'}$. Finally, the lifting-suppression domain-relevant feature ${F_{1LS}^{-}}$ is obtained by scaling the re-weighted ${F_{1UR}^{-}}$ via  $Q_{11}^{'}$ computed as follows \begin{equation}
F_{1LS}^{-}=\beta\times  reshape(Q_{11}^{'}\otimes reshape(F_{1UR}^{-}))+F_{1UR}^{-}
\end{equation}where $\otimes$ is matrix multiplication, $\beta$ is a learnable scale parameter and $reshape(\cdot)$ is used to match dimension. Without loss of generality, ${F_{2LS}^{-}}$, ${F_{3LS}^{-}}$, ${F_{4LS}^{-}}$ w.r.t. ${F_{2UR}^{-}}$, ${F_{3UR}^{-}}$, ${F_{4UR}^{-}}$ from multiple DDMs can be obtained through similar pipelines as above. Then, we sum these 4 kinds of features to obtain purified domain-relevant feature $F^{-}$.

\subsubsection{Domain Expansion}
Inspired by ISDA~\cite{wang2019implicit}, an implicit semantic data augmentation model for image classification, in this paper, the source domain expansion is achieved by random feature transformation in an implicit way without increasing training burden. Specifically, by pooling ${F^{+}}$ and ${F^{-}}$, identity-related and domain-related feature ${f^{+}}$ and ${f^{-}}$ can be obtained, respectively. 
The domain-wise co-variance matrix $\Sigma_{j}$ of domain $D_j$ is calculated through the domain-wise features ${f^{j}}^{-}$. Then we sample $K$ random vectors $\xi^{j}_{k}$ from domain-wise zero-mean multivariate normal distribution $\mathcal{N}\left(0, \Sigma_{j}\right)$ as the potential transformation directions, where $k\in \{1, \ldots, K\}$. 
For an identity-relevant feature ${f_{i}^{n}}^{+}$ of the $i$-th sample from domain $D_n$, its $K$ new augmented features $\left\{\left({{f}_{i1}^{nj}}^{+}, \ldots, {{f}_{iK}^{nj}}^{+}), y_{i}^{n}\right)\right\}_{i=1}^{N_{n}}$ could be obtained by exploiting the domain statistics of domain $D_{j\neq n}$ via the following feature expansion method, 
 \begin{equation}\label{eq1}{{{f}}_{ik}^{nj}}^{+} = {{f}_{i}^{n}}^{+} + \xi_{k}^{j}\vspace{-0.1cm}\end{equation} where $\xi_{k}^{j}\sim\mathcal{N}\left(0, \Sigma_{j}\right)$ and $k\in \{1, \ldots, K\}$.
 Then, for totally $S$ domains, the cross-entropy loss after $K$-times feature expansion for identity classification can be written as 
 \begin{equation}\label{eq2}
\mathcal{L}_{CE}^K=\sum_{n=1}^{S}\frac{1}{N_{n}} \sum_{i=1}^{N_{n}} \frac{1}{K} \sum_{k=1}^{K} CE\;({{f}_{i}^{n}}^{+} + \xi_{k}^{j})
\vspace{-0.1cm} \end{equation} where $CE( \cdot )$ is the cross-entropy loss, $N_{n}$ is the sample number of domain $D_n$.

However, generating $K$ new features and applying supervised loss on them would result in low computational efficiency when $K$ is large, and the value of $K$ is difficult to determine. In this paper, we suppose $K\to\infty$, 
then the supervised training on an infinite number of samples of Eq.~\ref{eq2} can be implemented efficiently in term of Taylor expansion. Specifically the above loss can be derived as\begin{equation}\label{eq3}
\begin{aligned}
\mathcal{L}_{CE}^{K \to \infty}\!\!  = \!\!\sum_{n=1}^{S}\frac{1}{N_{n}} \sum_{i=1}^{N_{n}} & \frac{1}{K} \sum_{k=1}^{K} ({CE}\;({{f}_{i}^{n}}^{+})\!\\& + \!(\nabla{CE}\;({{f}_{i}^{n}}^{+}))^T \xi_{k}^{j} \\
 &+\!\frac{1}{2} (\xi_{k}^{j})^T \; \! \nabla^2{CE}({{f}_{i}^{n}}^{+})\;\xi_{k}^{j}+ R)
\end{aligned}
\end{equation}
where $\nabla{CE}( \cdot )$ denotes the first-order derivative of the cross-entropy loss, and $\nabla^2{CE}( \cdot )$ denotes the second-order derivative. $R$ is the higher order term in Taylor Formula, which is empirically omitted in the following derivation due to its nature of infinitesimal. We observe that by excluding the omitted $R$ term, Eq. (\ref{eq3}) mainly consists of three parts. We deduce each item as follows.

The first item is the commonly used cross-entropy loss. The second item can be written as\begin{equation} \small \label{eq4} \mathcal{L}_{CE\_2nd}^{K \to \infty}=\sum_{n=1}^{S}\frac{1}{N_{n}} \sum_{i=1}^{N_{n}}(\nabla{CE}\;({{f}_{i}^{n}}^{+}))^T  \frac{1}{K} \sum_{k=1}^{K}\xi_{k}^{j}=0
\end{equation} Clearly, when $K \rightarrow \infty$, Eq. (\ref{eq4}) actually introduces the expectation of zero-mean gaussian variables according to the law of large numbers. Therefore the second item in Eq. (\ref{eq3}), namely the Eq. (\ref{eq4}), becomes zero.

As for the third item in Eq. (\ref{eq3}), we can derive it as\begin{equation} \small \label{eq8}
\begin{aligned}
\mathcal{L}_{CE\_3rd}^{K \to \infty} &= \!\!\sum_{n=1}^{S}\frac{1}{N_{n}} \sum_{i=1}^{N_{n}}  \frac{1}{K} \sum_{k=1}^{K}\frac{1}{2} (\xi_{k}^{j})^T \; \nabla^2{CE}({{f}_{i}^{n}}^{+})\;\xi_{k}^{j} \\
&=\!\!\sum_{n=1}^{S}\frac{1}{2N_{n}} \sum_{i=1}^{N_{n}}  \frac{1}{K} \sum_{k=1}^{K}tr(\xi_{k}^{j}\; \!\! (\xi_{k}^{j})^T \; \!\nabla^2{CE}({{f}_{i}^{n}}^{+})) \\
&=\!\!\sum_{n=1}^{S}\frac{1}{2N_{n}}\!\! \sum_{i=1}^{N_{n}} \!tr (( \frac{1}{K}\sum_{k=1}^{K} \xi_{k}^{j}\;\!\! (\xi_{k}^{j})^T )\nabla^2{CE}({{f}_{i}^{n}}^{+}))  \\
&=\!\!\sum_{n=1}^{S}\frac{1}{2N_{n}} \sum_{i=1}^{N_{n}} tr(\Sigma_{j}\nabla^2{CE}({{f}_{i}^{n}}^{+}))
\end{aligned}
\end{equation}
where tr$( \cdot )$ is the trace of a matrix.

\begin{algorithm}
\caption{The training process of LDE}\label{algo1}
\begin{algorithmic}[1]
\Require $S$ source domains $\left\{{D}_{1}, \ldots, {D}_{S}\right\}$, where ${D}_{j}=\small{\left\{\left(\boldsymbol{x}_{i}^{j}, y_{i}^{j}\right)\right\}_{i=1}^{N_{j}}}$, and ${N}_{j}$ is the number of images in domain ${D}_{j}$.
\Ensure Parameters of entire network $\Theta $.
\State {\textbf{Initialization}. Iteration counter $n$ = 0. Total iteration $m$. Initialize $\Theta $ with parameters pre-trained on ImageNet. Ensure trade-off hyper-parameter $\lambda$.}
\While{$n < m$}
\State Sampling $\left(\boldsymbol{x}_{i}^{j}, y_{i}^{j}\right)$ from $\left\{{D}_{1}, \ldots, {D}_{S}\right\}$.
\State \textbf{Step 1: Domain Decoupling.}

 $\bullet$ Get identity-related features ${f^{+}}$ through a dual-stream network with DDMs and MSLS.

 $\bullet$ Compute the cross-entropy loss $CE({{f}}^{+})$ by ${f^{+}}$, get the second-order derivative $\nabla^2{CE}({{f}}^{+})$.
\State \textbf{Step 2: Domain Expansion.}

$\bullet$ Compute the $L_{\infty}$-cross-entropy loss using Eq. (\ref{eq9}).

$\bullet$ Compute the total loss using Eq. (\ref{eq10}).

$\bullet$ Compute the gradient of Eq. (\ref{eq10}), and update $\Theta $.
\State $n = n + 1$.
\EndWhile
\end{algorithmic}
\end{algorithm}

By substituting Eq. (\ref{eq8}) into Eq. (\ref{eq3}), the $L_{\infty}$-cross-entropy loss is finalized as\begin{equation} \begin{aligned} \label{eq9} \mathcal{L}_{CE}^{K \to \infty}\!\!&=\!\!\sum_{n=1}^{S}\!\!\frac{1}{N_{n}} \!\sum_{i=1}^{N_{n}}\!({CE}\;({{f}_{i}^{n}}^{+}) \\&+\frac{\lambda}{2}\!\sum_{j=1}^{S}\!tr(\Sigma_j\nabla^2{CE}({{f}_{i}^{n}}^{+})))
\end{aligned} \end{equation} where $\lambda$ is a hyper-parameter to control the strength of reciprocal domain breadth expansion. When $\lambda = 0$, $\mathcal{L}_{CE}^{K\to\infty}$ will be degenerated into standard cross-entropy loss. The goal of Domain Expansion could be achieved latently by optimizing Eq. (\ref{eq9}) on the original features.

In order to facilitate the intra-class similarity and inter-class dissimilarity learning, the triplet loss $\mathcal{L}_{\mathrm{Tri}}$~\cite{hermans2017defense} is utilized. The overall objective for person ReID is written as \begin{equation} \label{eq10} \mathcal{L}_{\mathrm{All}}=\mathcal{L}_{\mathrm{Tri}}+\mathcal{L}_{CE}^{K\to \infty}
\end{equation}

The algorithm details of the proposed LDE are summarized in Algorithm~\ref{algo1}.
\section{Experiments}
\subsection{Datasets}

\textbf{1) Datasets for evaluating OWD}. To demonstrate the potential of OWD to facilitate model transferability and generalization, more than 10 widely-used datasets are adopted as source domains for comparisons. There are three different types of benchmarks: real-world data, web data, and synthetic data. Real-world data includes large-scale CUHK03 \cite{6909421}, Market1501 \cite{zheng2015scalable}, Dukemtmc \cite{zheng2017unlabeled}, and MSMT17 \cite{8578114}. For web benchmark, LaST \cite{shu2021large} collected under cloth-changing situations is exploited as a large-scale benchmark. Additionally, six synthetic datasets, SOMAset \cite{Barbosa2018Looking}, SyRI \cite{bak2018domain}, PersonX \cite{sun2019dissecting}, RandPerson \cite{wang2020surpassing}, UnrealPerson \cite{zhang2021unrealperson}, and ClonedPerson \cite{wang2022cloning} are also used as source domain respectively for comparisons.

\textbf{2) Datasets for evaluating LDE}. To comprehensively evaluate the performance of LDE, we conduct domain generalization (DG) experiments, where the models can only access labeled source domains for training. First, 
PRID \cite{Hirzer2011PersonRB}, GRID \cite{2014Person}, iLIDS \cite{2009Associating}, and VIPeR \cite{2008Viewpoint} participate in the evaluation as four small-scale target domains. Their source domains are a mixture of multi-dataset benchmarks, containing CUHK02, CUHK03, Market1501, Dukemtmc, and CUHK-SYSU PersonSearch \cite{xiao2017joint}. 
Second, we use Market1501 and Dukemtmc as two large-scale target domains respectively. Their corresponding source domains are the data of CUHK03+Dukemtmc+MSMT17 and Market1501+MSMT17+CUHK03, respectively. Besides, we evaluate the proposed LDE on the collected OWD under different settings, including multi-source DG, traditional person ReID, and single-source DG. Specifically, multi-source DG is from Market1501 + CUHK03 + MSMT17 datasets to the collected OWD, traditional person ReID exploits the open-scene protocol on OWD, and single-source DG is from the collected OWD to Market1501, Dukemtmc, and MSMT17, respectively.

\subsection{Implementation Details}
\textbf{1) Setting in evaluating OWD}. To verify the generalization ability of OWD, we conduct direct transfer experiments based on the ResNet50 backbone \cite{7780459} with softmax loss following the setting in RandPerson \cite{wang2020surpassing}. This setting is followed in the majority of experiments except for the situations where UnrealPerson and ClonedPerson are used as source domains respectively. Specifically, we deploy BoT \cite{Luo_2019_CVPR_Workshops} and QAConv 2.0 \cite{ liao2022graph} on OWD for comparison with UnrealPerson and ClonedPerson, respectively.

\textbf{2) Setting in evaluating LDE}. Similar to the strategy in \cite{2021Person30K}, we adopt the model of FastReID \cite{he2020fastreid} as the baseline of our LDE, and the comparisons between models also follow this setting to ensure the fairness in comparison. During training, random horizontal flipping is used for data augmentation. The batch size is set to 64. The learning rate starts from $3.5 \times 10^{-4}$ and decays to $7.7 \times 10^{-7}$. We optimize the model with Adam optimizer. The hyper-parameter $\lambda$ is set to 1.0. We share the OWD benchmark, LDE model and codes with the community.

\subsection{Evaluation Results on OWD}\label{sec5.2}

\begin{table*}[t]
\caption{The performance (\%) of some existing advanced methods on OWD dataset under different evaluation protocols}\label{per_owd}
\begin{tabular*}{1.0\textwidth}{@{\extracolsep\fill}lccccccccc}
\toprule%
& \multicolumn{3}{@{}c@{}}{close-scene} & \multicolumn{3}{@{}c@{}}{open-scene} & \multicolumn{3}{@{}c@{}}{day-night} \\\cmidrule{2-4}\cmidrule{5-7}\cmidrule{8-10}%
Methods & Rank-1 & Rank-5 & mAP &  Rank-1 & Rank-5  & mAP &Rank-1 & Rank-5  & mAP \\
\midrule
BoT  & 79.2 & 90.3 & 60.3 & 69.7 & 83.0  & 49.0 & 54.5 & 70.3 & 33.0 \\
IBN  & 83.4 & 92.7  & 65.9 & 75.2 & 86.7 & 54.5 & 63.6 & 78.2  & 40.3 \\
AGW  & 83.7 & 92.8 & 66.8 & 74.6 & 85.9  & 53.7 & 61.4 & 76.0 & 38.6 \\
MGN  & 85.9 & 93.8 & 70.2 & 75.6 & 86.5 & 56.1 & 64.2 & 79.2 & 42.7 \\
SBS & 86.6 & 94.0 & 70.1 & 79.4 & 88.9 & 59.6 & 69.5 & 82.3 & 45.5 \\
\hline
\end{tabular*}
\end{table*}

\textbf{1) General Evaluation.} As presented in section \ref{sect_eval}, OWD provides three evaluation protocols based on different domain gaps. Table \ref{per_owd} shows the comparison results on the proposed OWD dataset, by using classical BoT \cite{Luo_2019_CVPR_Workshops}, BoT with IBN module \cite{pan2018two}, AGW \cite{arxiv20reidsurvey}, MGN \cite{2018Learning} and SBS \cite{he2020fastreid}. Although these models have performed well on previous benchmarks, their performances on OWD are unsatisfactory. For example, SBS achieves 95.4\% Rank-1 accuracy and 88.2\% mAP on the Market1501 dataset, but only 69.5\% Rank-1 and 45.5\% mAP on OWD under the \textbf{day-night} protocol. These results indicate that the open-world and challenging scenes introduced by OWD can not be conquered by these classical methods. The performance drop of the model from \textbf{close-scene} to \textbf{open-scene} proves that dynamic wild variations such as lighting, clothing, backgrounds, adverse conditions etc. increase the difficulty of pedestrian retrieval. Notably, the performances drop dramatically under the \textbf{day-night} setting, which proves that in practical application, the model trained on the daytime pedestrian image is difficult to adapt to the night environment with poor lighting conditions. The \textbf{day-night} protocol of OWD provides a new benchmark for this tough scenario to promote the development of the ReID community in this regard. Therefore, OWD can be used to comprehensively evaluate 
the robustness and generalization of a model for instructing practical deployment. 

\begin{table*}[t]
\caption{Performance (\%) across different datasets (Rank-1/mAP), where '\dag` and '\ddag` represent the same models as UnrealPerson and ClonedPerson are used for comparison, i.e., the BoT \cite{Luo_2019_CVPR_Workshops} for UnrealPerson and the QAConv 2.0 \cite{ liao2022graph} for ClonedPerson. The best results are in bold}\label{crossdata}%
\setlength\tabcolsep{0.5pt}

\begin{tabular*}{\textwidth}{@{\extracolsep\fill}lccccc}
\toprule
Source Dataset & Source Style & CUHK03 & Market1501 & Dukemtmc & MSMT17 \\
\midrule
SOMAset \cite{Barbosa2018Looking} & Synthetic data & 0.4 / 0.4 &4.5 / 1.3 & 4.0 / 1.0 & 1.4 / 0.3  \\
SyRI \cite{bak2018domain}& Synthetic data & 4.1 / 3.5 & 4.5 / 1.3 & 23.7 / 9.0 & 16.4 / 4.4 \\
PersonX \cite{sun2019dissecting} & Synthetic data& 7.4 / 6.2 &44.0 / 20.4 & 35.4 / 18.1 & 11.7 / 3.6  \\
RandPerson \cite{wang2020surpassing}& Synthetic data & \textbf{13.4} / 10.8 & 55.6 / 28.8 & 47.6 / 27.1 & 20.1 / 6.3  \\
CUHK03 \cite{6909421}& Real-world data & - &32.7 / 12.8 & 25.5 / 11.0 & 12.5 / 3.0 \\
Market1501 \cite{zheng2015scalable} & Real-world data& 5.6 / 5.0 &- & 34.3 / 17.9 & 12.1 / 3.4 \\
Dukemtmc \cite{zheng2017unlabeled}& Real-world data & 7.0 / 6.1 & 46.5 / 19.6 & - & 20.2 / 6.0 \\

MSMT17 \cite{8578114} & Real-world data& 11.2 / 10.3 & 51.8 / 24.1 & 55.7 / 32.9 & - \\
LaST \cite{shu2021large} &Web data& 4.4 / 4.0 & 46.6 / 22.0 & 43.7 / 25.0 & 8.8 / 2.8 \\
CUHK03+MSMT17+Dukemtmc & Real-world data& - & \textbf{69.1} / 38.5 & - & - \\
CUHK03+Market1501+Dukemtmc & Real-world data& - & - & - & 23.6 / 7.7 \\
OWD without nighttime data & Real-world data& 12.1 / 10.3 & 64.3 / 35.8 & 53.4 / 31.8 & 20.5 / 7.5\\
OWD& Real-world data & 13.2 / \textbf{11.7} & 68.3 / \textbf{39.8} & \textbf{56.8 / 34.4} & \textbf{32.5 / 11.5} \\
UnrealPerson\dag \cite{zhang2021unrealperson} & Synthetic data&- & 64.0 / 37.2 & 58.0 / 37.5 & 26.8 / 9.9  \\
OWD\dag & Real-world data &- & \textbf{80.9 / 57.8} & \textbf{66.2 / 47.9} & \textbf{35.4 / 18.3}  \\
ClonedPerson\ddag \cite{wang2022cloning}& Synthetic data & 22.6 / 21.8 & 84.5 / 59.9 & - & 49.1 / 18.5  \\
OWD\ddag  & Real-world data& \textbf{27.5 / 27.5} & \textbf{87.8 / 68.3} & - & \textbf{66.0 / 34.0}  \\
\hline
\end{tabular*}
\end{table*}

\begin{table*}[t]
\caption{Performance (\%) on close-scene datasets (Market1501 and Dukemtmc) when the model is pre-trained on different benchmarks}\label{pre1}
\small
\begin{tabular*}{\textwidth}{@{\extracolsep\fill}lcccc}
\toprule%
& \multicolumn{2}{c@{}}{Market1501} & \multicolumn{2}{c@{}}{DukeMTMC} \\\cmidrule{2-3}\cmidrule{4-5}%
Pre-Training  & Rank-1 & mAP & Rank-1 & mAP \\
\midrule
ImageNet \cite{russakovsky2015imagenet} & 89.0 & 71.1 & 78.7 & 62.8  \\
MSMT17 \cite{8578114}  & 92.5 & 79.2  & 84.4 & 70.7  \\
LaST \cite{shu2021large}  & 91.4 & 79.1  & 82.5 & 69.0  \\
LaST\underline{ }Cloth \cite{shu2021large}  & 93.1 & 81.7  & 84.5 & 71.7  \\
OWD & \textbf{95.4} & \textbf{88.4}  & \textbf{87.8} & \textbf{76.9}  \\
\hline
\end{tabular*}
\end{table*}

\textbf{2) Cross-Dataset Evaluation.} To verify the generalization potential of OWD, following the setting in RandPerson \cite{wang2020surpassing}, we train a model on one dataset and test it on another to conduct the open-domain transferability evaluation. As shown in Table \ref{crossdata}, different person datasets, including real-world data, web data and synthetic data, are used as source data for training. We observe that OWD has a stronger transfer ability compared with these commonly used datasets in most cases. For example, compared with MSMT17 that has a similar size, OWD improves significantly, when tested on another open-domain dataset, i.e., Market1501. In addition, for multiple sources domain generalization (i.e., CUHK03+Market1501+Dukemtmc in Table \ref{crossdata}), it can effectively improve the transferability (23.6\% Rank-1 accuracy), but still inferior to our OWD (32.5\% Rank-1 accuracy), when tested on MSMT17. For the combination CUHK03+MSMT17+Dukemtmc with a larger scale, OWD could provide competitive transfer performance when tested on Market1501. Also, although LaST \cite{shu2021large} provides larger-scale pedestrian data, its transferability is insufficient with only 2.8\% mAP when tested on MSMT17. This may be related to the sampling manner of LaST, that is, cropping images from web videos. To ensure the broadcast effect, web videos and movies are usually clear and delicate, which circumvents the tough scenes in applications and limits the diversity of pedestrian samples.

Furthermore, the nighttime samples are challenging due to insufficient lighting conditions. In order to verify whether the nighttime data is beneficial to the generalization ability of models, we remove the nighttime data in OWD and re-evaluate on open domains in Table \ref{crossdata}. The results demonstrate that without nighttime data, 
the Rank-1 accuracy decreases by 12.0$\%$ from 32.5$\%$ to 20.5$\%$ on MSMT17, which indicate that nighttime data is helpful to improve generalization of models and the necessity of OWD.

Besides, compared with the large-scale synthetic datasets such as RandPerson \cite{wang2020surpassing}, UnrealPerson \cite{zhang2021unrealperson}, and  ClonedPerson \cite{wang2022cloning}, OWD could also perform much better. In contrast, the proposed OWD can provide stronger transfer capability (i.e., OWD\dag) when the same model BoT~\cite{Luo_2019_CVPR_Workshops} as UnrealPerson\dag is used. In addition, when a more robust model, i.e. QAConv 2.0 \cite{liao2022graph} used by ClonedPerson\ddag, is deployed, OWD shows a stronger generalization capability (i.e., OWD\ddag). The is due to that although these synthetic datasets provide a large number of pedestrian samples, significant domain gaps prevent them from being generalized to real scenes.

\begin{table*}[t]
\caption{Performance (\%) of combined training datasets. The model is trained on a target training data (e.g., train set of Market1501) plus an extra benchmark, and then tested on the target test data}\label{combine}
\small
\begin{tabular*}{\textwidth}{@{\extracolsep\fill}lcccc}
\toprule%
& \multicolumn{2}{c@{}}{Market1501} & \multicolumn{2}{c@{}}{DukeMTMC} \\\cmidrule{2-3}\cmidrule{4-5}%
Training Set  & Rank-1 & mAP & Rank-1 & mAP \\
\midrule
+MSMT17 \cite{8578114}  & 91.7 & 77.4  & 83.4 & 69.7  \\
+LaST\underline{ }Cloth \cite{shu2021large}  & 92.3 & 80.0  & 84.3 & 70.2  \\
+OWD & \textbf{94.5} & \textbf{84.5}  & \textbf{85.4} & \textbf{72.3}  \\
\hline
\end{tabular*}
\end{table*}

\textbf{3) Pre-trained performance on close-scene datasets.} At present, most ReID models are pre-trained through ImageNet \cite{russakovsky2015imagenet}, a large-scale image classification benchmark. Although it can provide a general feature representation for pedestrian retrieval tasks, the robustness is still limited. It is valuable if a Person based ImageNet exists. Therefore, we explore the performance by pre-training on different person datasets. That is, the ReID model is first pre-trained on a specific person benchmark from scratch, then trained and tested on a close-scene dataset, i.e., Market1501 and DukeMTMC in this experiment. In the experiments, ImageNet, large-scale MSMT17, LaST, and proposed OWD are exploited as the pre-training benchmark respectively for comparison.

Table \ref{pre1} illustrates the comparison results of different benchmarks used for pre-training. Compared with the general ImageNet, the model can achieve better performance by pre-training on task-specific pedestrian data, such as MSMT17, LaST and OWD. As can be observed, the proposed OWD can provide a more robust pre-training model for down-stream tasks. Specifically, the model pre-trained on OWD achieves 88.4\% and 76.9\% mAP on Market1501 and Dukemtmc datasets, respectively, which are much better than the largest LaST\_Cloth benchmark (81.7\% and 71.7\% mAP). 

\begin{table*}[t]
\caption{Direct transfer experiments under the cloth-changing situation. The ReID model is trained on one specific benchmark and then tested on target cloth-changing data directly. The best results are in bold}\label{cloth1}
\small
\begin{tabular*}{\textwidth}{@{\extracolsep\fill}lcccccccc}
\toprule%
& \multicolumn{2}{c@{}}{PRCC} & \multicolumn{2}{c@{}}{Celeb-reID} & \multicolumn{2}{c@{}}{LTCC} & \multicolumn{2}{c@{}}{DeepChange} \\\cmidrule{2-3}\cmidrule{4-5}\cmidrule{6-7}\cmidrule{8-9}%
Training Set  & Rank-1 & mAP & Rank-1 & mAP & Rank-1 & mAP & Rank-1 & mAP \\
\midrule
ImageNet \cite{russakovsky2015imagenet} & 24.7 & 13.5 & 28.7 & 3.0 & 20.7 & 4.4 & 4.2 & 1.0 \\
Market1501 \cite{zheng2015scalable}  & 29.0 & 24.3 & 36.7 & 3.7 & 30.7 & 8.5 & 12.6 & 4.3  \\
DukeMTMC \cite{zheng2017unlabeled}  & 28.3 & 24.1  & 40.9 & 4.6 & 29.0 & 9.4 & 15.8 & 5.0 \\
MSMT17 \cite{8578114}  & 26.2 & 24.6  & 43.4 & 5.0 & 33.8 & 12.0 & 18.3 & 6.4 \\
LaST \cite{shu2021large} & \textbf{39.3} & 32.6  & 47.0 & \textbf{7.0} & 33.8 & 9.3 & 8.8 & 2.8 \\
OWD & 34.7 & \textbf{35.3} & \textbf{47.2} & 6.1 & \textbf{41.8} & \textbf{16.3} & \textbf{24.9} & \textbf{9.9}\\
\hline
\end{tabular*}
\end{table*}

\textbf{4) Performance of combined datasets.} To further verify the effectiveness of the proposed OWD improving others, we combine different ReID benchmark with a target dataset for joint training. The experimental results are shown in Table \ref{combine}. Take Market1501 as an example, when its training data is mixed with others, the model can achieve better performance on its test set. The results show that after combining OWD (i.e., +OWD), the best performance can be achieved than +MSMT17 and +LaST\_Cloth. The best is also achieved on DukeMTMC. This demonstrates that OWD can work as a stronger auxiliary benchmark for improving a scene-specific dataset and the model.

\textbf{5) Direct transfer evaluation on cloth-changing setting.} In practical applications, 
the cloth-changing setting is a challenging but practical issue. In this case, a ReID model needs to rely on the morphological characteristics of the pedestrian for identification. At present, quite a lot of work focus on this problem, and specific benchmarks are developed to promote the development of this issue, such as PRCC \cite{yang2019person}, Celeb-reID \cite{huang2019beyond}, LaST \cite{shu2021large}, LTCC \cite{qian2020long}, and Deepchange \cite{xu2023deepchange}.

To further evaluate the generalization capacity of OWD, direct transfer experiments were conducted on cloth-changing benchmarks, i.e., training a ReID model on a benchmark, then deploying it on a cloth-changing target domain without fine-tuning or re-training. Specifically, six large-scale datasets are used as source domains to train the model respectively and then tested directly on cloth-changing PRCC and Celeb-reID datasets. The results are shown in Table \ref{cloth1}. We observe that although the scale of ImageNet is far greater than other person datasets, its transfer performance is quite inferior to the pedestrian-based benchmarks. This is due to most of the prior information provided by ImageNet are non-human clues. Although these clues are universal, they are still insufficient when facing specific ReID task. Compared with the traditional pedestrian-based close-scene benchmarks, OWD provides comparable transfer performance with the mAP of 35.3\% on PRCC and 10.7\% higher than MSMT17. It is worth noting that compared with LaST, a large-scale clothing-changing benchmark, OWD can provide a more comprehensive generalization capability. Furthermore, on more realistic scenes, such as DeepChange and LTCC benchmarks, OWD could also demonstrate better performance. For example, the Rank-1 accuracy is 6.6\% higher than large-scale MSMT17 on DeepChange. The experimental results demonstrate the generalization potential of OWD in practical and complex scenarios.   

\begin{table*}
\centering
\renewcommand{\arraystretch}{0.8}
\setlength\tabcolsep{3.8pt}
\footnotesize
\caption{Performance (\%) comparison with some SOTA DG person ReID methods. The best results are bolded}\label{perforcom}
\begin{tabular}{lcccccccccccccccccc}
\toprule%
& \multicolumn{2}{@{}c}{Average}& \multicolumn{4}{@{}@{}c@{}}{PRID} & \multicolumn{4}{@{}@{}c@{}}{GRID} & \multicolumn{4}{@{}@{}c@{}}{i-LIDS} & \multicolumn{4}{@{}@{}c@{}}{VIPeR}\\\cmidrule{2-3}\cmidrule{4-7}\cmidrule{8-11}\cmidrule{12-15}\cmidrule{16-19}%
Methods & R-1 & mAP & R-1 & R-5 &R-10 & mAP & R-1 & R-5 &R-10 & mAP & R-1 & R-5 &R-10 & mAP & R-1 & R-5 &R-10 & mAP  \\
\midrule
Agg\underline{ }PCB \cite{sun2018models} &40.6 & 49.0 &21.5 &42.6 &49.7 &32.0 &36.0 &53.7 &63.3 &44.7 &66.7 &81.7 &86.8 &73.9 &38.1 &53.2 &59.3 &45.4 \\
Agg\underline{ }Align \cite{zhang2017alignedreid} & 34.9 & 44.5  &17.2 &33.4 &39.6 &25.5 &15.9 &33.5 &41.4 &24.7 &63.8 &89.2 &95.5 &74.7 &42.8 &63.7 &73.6 &52.9\\
PPA \cite{Act2Param} & 42.1 & 52.6  &31.9 &61.1 &70.5 &45.3 &26.9 &50.5 &61.5 &38.0 &64.5 &83.8 &88.0 &72.7 &45.1 &65.1 &72.7 &54.5\\
Reptile \cite{nichol2018firstorder} & 28.1 & 37.1  &17.9 &33.8 &44.1 &26.9 &16.2 &29.4 &38.4 &23.0 &56.0 &80.7 &89.8 &67.1 &22.1 &39.4 &49.2 &31.3 \\
CrossGrad \cite{shankar2018generalizing} & 24.6 & 34.0  &18.8 &35.3 &46.0 &28.2  &9.0 &22.1 &30.1 &16.0 &49.7 &74.2 &83.8 &61.3 &20.9 &39.1 &49.7 &30.4 \\
MLDG \cite{li2017learning} & 29.3 & 39.4  &24.0 &48.0 &53.6 &35.4 &15.8 &31.1 &39.8 &23.6 &53.8 &78.7 &88.0 &65.2 &23.5 &43.8 &52.5 &33.5\\
DIMN \cite{2019Generalizable}  & 47.5 & 57.9 &39.2 &67.0 &76.7 &52.0 & 29.3 & 53.3 & 65.8 & 41.1 & 70.2 &  89.7  & 94.5 & 78.4  &51.2 &70.2 &76.0 &60.1\\
AugMining \cite{tamura2019augmented} &51.8 &- &34.3 &56.2 &65.7 &- & 46.6 & 67.5  & 76.1 & - & 76.3 &  93.0  &95.3 & - &49.8 &70.8 &77.0 &- \\
Switchable (BN+IN) \cite{luo2019switchable} &57.0 &65.6  &59.6 &78.6 &90.1 &69.4 & 39.3 &  58.8 &  68.1 &  48.1 & 77.3 & 91.2  & 94.8 & 83.5 &51.6 &72.9 &80.8 &61.4\\
DualNorm \cite{jia2019frustratingly} &62.7 &- &69.6 &- &- &- &43.7 &- &- &- &78.2 &- &- &- &59.4 &- &- &- \\
BoT \cite{Luo_2019_CVPR_Workshops} & 53.7 & 62.2  & 51.4 &- &-& 61.3 & 40.5&- &- & 49.6 & 74.7 &- &- & 81.3 & 48.2 &- &-& 56.7 \\
DDAN \cite{chen2021dual} &59.0 &63.1 &54.5 &62.7 &74.9 &58.9 & 50.6 & 62.1 & 73.8 & 55.7 & 78.5 & 85.3  & 92.5 & 81.5 &52.3 &60.6 &71.8 &56.4 \\
DDAN \cite{chen2021dual} w/ \cite{jia2019frustratingly} &60.9 &65.1 &62.9 &74.2 &85.3 &67.5 & 46.2 & 55.4 & 68.0 & 50.9 & 78.0 &  85.7  &   93.2 & 81.2 &56.5 &65.6 &76.3 &60.8 \\
MixStyle \cite{zhou2021domain} & 54.1 & 62.1  & 53.0 & - & -& 62.4 & 39.2& - & - & 46.8 & 75.0 & - & -& 81.4 & 49.4 & - & -& 57.9 \\
SNR \cite{jin2020style} &57.3 &66.4 &52.1 &- &- &66.5 &40.2 &- &- &47.7 &84.1 &- &- &89.9 &52.9 &- &- &61.3 \\
Sampler + GradDrop \cite{zhao2022revisiting} &66.6 &74.4 &\textbf{74.4} &\textbf{91.7} &\textbf{96.4} &\textbf{81.6} &49.9 &70.0 &78.6 &59.6 &80.7 &\textbf{95.7} &97.7 &87.2 &61.5 &77.3 &82.9 &69.2
\\
DMG-Net \cite{2021Person30K}  & 61.2 & 67.3 & 60.6 & - & -& 68.4 & 51.0& - & - & 56.6 & 79.3 & - & - & 83.9 & 53.9 & - & -& 60.4 \\
RaMoE \cite{dai2021generalizable} &61.5 & 69.1  &57.7 & -  & - & 67.3 &46.8 & -  & - & 54.2 &\textbf{85.0} & -  & - & \textbf{90.2} &56.6 & -  & - & 64.6\\
MetaBIN \cite{choi2021meta} &66.0 &73.6  &74.2 &89.7 &92.2 &81.0 &48.4 &70.3 &77.2 &57.9 &81.3 &95.0 &97.0 &87.0 &59.9 &78.4 &82.8 &68.6\\
MDA \cite{9880010} &\textbf{68.4} &73.0  &- &- &- &- &\textbf{61.2} &\textbf{83.4} &\textbf{88.9} &62.9 &80.4 &92.2 &95.0 &84.4 &63.5 &\textbf{80.6} &\textbf{84.2} &\textbf{71.7}\\
SHS+DSM \cite{Li2023StyleControllableGP} &67.3 &75.2  &71.0 &- &- &78.9 &50.7 &- &- &60.4 &80.7 &- &- &86.9 &\textbf{66.8} &- &- &74.4\\
LDE  & 67.4 & \textbf{75.2} & 72.0 & 91.0 & 93.0 & 79.1 & 58.4 & 79.2 & 86.4 & \textbf{67.7} & 83.3 & 95.0 & \textbf{97.8} & 88.3 & 55.7 & 76.6 & 83.5 & 65.7 \\
LDE with OWD  & 69.0 & 77.0 & 70.0 & 89.0 & 93.0 & 78.9 & 61.6 & 78.4 & 84.8 & 69.4 & 85.0 & 96.7 & 96.7 & 90.1 & 59.2 & 81.0 & 86.7 & 69.0 \\
\hline
\end{tabular}
\end{table*}

\subsection{Evaluation Results of LDE}

\textbf{1) Performance on Small-Scale Target Domains.} We show the performance comparison on four small-scale target domains of the proposed LDE with other competitive methods in Table \ref{perforcom}, including domain aggregation models (Agg\_PCB \cite{sun2018models}, Agg\_Align \cite{zhang2017alignedreid} and RaMoE \cite{dai2021generalizable}), general DG methods (CrossGrad \cite{shankar2018generalizing} and MLDG \cite{li2017learning}), general meta-learning methods (PPA \cite{Act2Param} and Reptile \cite{nichol2018firstorder}), DG person ReID methods (DIMN \cite{2019Generalizable}, DDAN \cite{chen2021dual}, DualNorm \cite{jia2019frustratingly}, DMG-Net \cite{2021Person30K}, MetaBIN \cite{choi2021meta}, MDA \cite{9880010}), and SHS + DSM \cite{Li2023StyleControllableGP}, and Sampler + GradDrop \cite{zhao2022revisiting}, and feature-level augmentation method MixStyle \cite{zhou2021domain}. To ensure fairness, we follow the experimental protocol in DMG-Net \cite{2021Person30K}. We can observe that the DG person ReID methods achieve much better performance compared with others. Comparing to DualNorm that employs IN to filter out style statistic variations and FN (Feature Normalization) to eliminate disparity in content statistics, LDE exploits normalization to capture domain-relevant features and achieves superior results on most datasets. Compared with the DMG-Net that designs a specific learning then generalization guided meta-learning procedure for DG person ReID, the proposed LDE achieves superior performance. MetaBIN generalizes normalization layers by simulating unsuccessful generalization scenarios beforehand in the meta-learning pipeline and obtains comparable results. MDA aligns distributions across source and target domains by designing a meta-learning strategy to simulate the real train-test process. However, meta-learning based DG methods usually introduce cumbersome training pipeline, while our method is easy to implement through a general dual-stream paradigm with domain decoupling. Different from RaMoE that exploits an effective voting-based mixture mechanism to dynamically leverage multiple source models, LDE resorts to only one model and provides encouraging results. SHS + DSM proposes an explicit dynamic style mixing method to increase the diversity of source domains, while LDE conducts latent data augmentation in feature space. Compared with these different types of methods, our LDE could effectively achieve competitive generalization performance via implicit domain expansion. Besides, we deploy the proposed LDE together with the collected OWD benchmark. We observe the generalization performance of the model is improved by 1.8\% in average mAP accuracy. This indicates that a more practical and diverse data benchmark helps the model extract invariant features.

\begin{table*}[t]
\caption{Performance (\%) comparison on large-scale target benchmarks}\label{largedata}
\small
\begin{tabular*}{\textwidth}{@{\extracolsep\fill}lcccc}
\toprule%
& \multicolumn{2}{@{}c}{Market1501} & \multicolumn{2}{@{}c}{Dukemtmc} \\\cmidrule{2-3}\cmidrule{4-5}%
Method & Rank-1 & mAP & Rank-1 & mAP\\
\midrule
QAConv$_{50}$ \cite{liao2020interpretable}& 68.6 & 39.5 & 64.9 & 43.4  \\
IBN-Net \cite{pan2018two} & 73.4 & 43.0 & 64.9 & 45.7  \\
MixStyle \cite{zhou2021domain} & 71.5 & 42.3  & 65.1 & 44.8  \\
CBN \cite{zhuang2020rethinking} & 74.7 & 47.3  & \textbf{70.0} & 50.1  \\
SNR \cite{jin2020style} & 75.2 & 48.5  & 66.7 & 48.3  \\
OSNet \cite{zhou2021learning} & 72.5 & 44.2  & 65.2 & 47.0  \\
OSNet-IBN \cite{zhou2021learning} & 73.0 & 44.9  & 64.6 & 45.7  \\
OSNet-AIN \cite{zhou2021learning} & 73.3 & 45.8  & 65.6 & 47.2  \\
M$^{3}$L \cite{zhao2021learning} & 74.5 & 48.1  & 69.4 & 50.5  \\
LDE & \textbf{78.4} & \textbf{52.8} & 68.9 & \textbf{51.5} \\
LDE with OWD & 82.8 & 58.5 & 74.6 & 56.9 \\
\hline
\end{tabular*}
\end{table*}

\begin{figure*}[t]
\includegraphics[width=1.0\linewidth,height=0.27\linewidth]{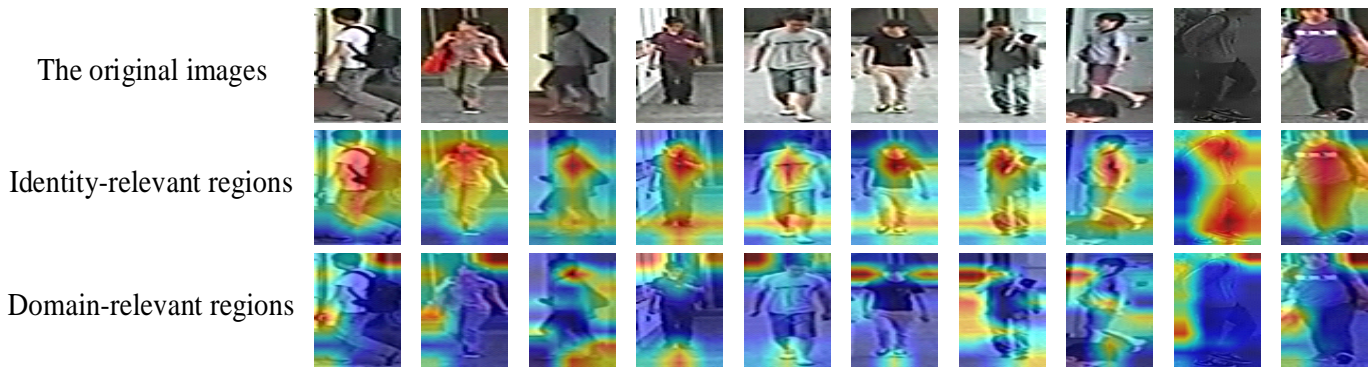}
   \caption{Visualization of heat maps via Grad-cam on identity-related features and domain-related features learned by LDE. The former focuses on discriminative regions for ReID, while the latter focuses on background regions, providing accurate domain style for domain expansion}
\label{attention}
\end{figure*}

\textbf{2) Performance on Large-Scale Target Domains.} We further evaluate the performance of LDE on large-scale target domains. Following QAConv$_{50}$ \cite{liao2020interpretable}, we use the commonly used Market1501 and Dukemtmc datasets as large-scale target domains. The source domains of Market1501 are the training data of Dukemtmc+Cuhk03+MSMT17, and the source domains of Dukemtmc are the training data of Market1501+Cuhk03+MSMT17. The results are presented in Table \ref{largedata}, from which we observe LDE achieves competitive performance comparing to other DG ReID models. 
For example, LDE achieves a 7.0\% improvement in mAP when tested on Market1501 compared with the advanced OSNet-AIN \cite{zhou2021learning}. We describe some representative models as follows. CBN \cite{zhuang2020rethinking} exploits normalization to eliminate the distribution gap of different cameras, thereby mitigating the domain difference, while LDE adopts standardized operation. SNR \cite{jin2020style} exploits normalization to filter out style variations, and then distills identity-relevant features from the removed information and restitutes it to the network, while LDE resorts to the attention mechanism to decouple identity-relevant features. M$^{3}$L \cite{zhao2021learning} adopts a meta-learning pipeline to improve the generalization ability of ReID, which may cost more computing resources, while LDE follows a general dual-stream paradigm. Compared with the above SOTA DG ReID methods, the proposed LDE achieves comparable performance on large-scale target domains in an easy-to-implement manner. 
Furthermore, we deploy the proposed LDE together with our developed OWD benchmark. We observe the generalization performance of the model is significantly improved by 5\% in both Rank-1 and mAP. For example, on Market1501, the mAP accuracy increases from 52.8\% to 58.5\% by introducing OWD. Therefore, for dynamic wild application, a powerful DG model and an open-world benchmark are equally crucial and necessary.
Although LDE is simply implemented with sample-level feature augmentation, it efficiently facilitates domain invariant learning if a dataset has no sufficient image-level diversity.

\begin{table*}[t]
\centering
\setlength\tabcolsep{8.2pt}
\small
\caption{Performance (\%) comparison with some SOTA DG person ReID methods on the OWD dataset, where `1' indicates multi-source domain generalization that is from Market1501 + CUHK03 + MSMT17 datasets to the collected OWD, `2' is traditional person ReID on OWD, and `3' represents single-source domain generalization that is from the collected OWD to Market1501, Dukemtmc, and MSMT17, respectively. The best results are bolded}\label{perform_owd}
\begin{tabular}{lcccccccccccccccccc}
\toprule%
& \multicolumn{2}{@{}c}{OWD\footnotemark[1]}& \multicolumn{2}{c@{}}{OWD\footnotemark[2]} & \multicolumn{2}{c@{}}{Market1501\footnotemark[3]} & \multicolumn{2}{c@{}}{Dukemtmc\footnotemark[3]} & \multicolumn{2}{c@{}}{MSMT17\footnotemark[3]}\\\cmidrule{2-3}\cmidrule{4-5}\cmidrule{6-7}\cmidrule{8-9}\cmidrule{10-11}%
Methods & R-1 & mAP & R-1 & mAP & R-1  & mAP & R-1 & mAP & R-1 & mAP  \\
\midrule
DDAN \cite{chen2021dual} & 41.2 & 25.7 & 53.0 & 37.1 & 78.1 & 51.5 & 54.0 & 32.0 & 24.0 & 8.8 \\
SNR \cite{jin2020style} & 48.6 & 36.2 & 63.5 & 51.7 & 80.6 & 57.4 & 63.9 & 44.6 & 38.1 & 18.9 \\
CBN \cite{zhuang2020rethinking} & 41.6 & 24.5 & 60.6 & 47.2 & 86.9 & 65.1 &68.4 & 48.4 & 36.5 & 16.5 \\
QAConv$_{50}$ \cite{liao2020interpretable} & 45.6 & 29.5 & 62.6 & 48.4 & 78.6 & 53.3 &64.3 & 41.7 & 35.6 & 15.2 \\
OSNet-AIN \cite{zhou2021learning} & 48.2 & 33.2 & 61.6 & 49.4 & 77.9 & 53.6 &60.1 & 40.3 & 30.1 & 13.4 \\
MetaBIN \cite{choi2021meta} & 52.4 & 38.5 & 62.9 & 50.5 & 82.5 & 59.8 &69.4 & 47.1 & 36.6 & 17.2  \\
QAConv-GS \cite{liao2022graph} & 50.5 & 38.7 & 53.8 & 41.6  & 82.3 & 55.2 &66.8 & 44.5 & \textbf{43.8} & 18.6  \\
LDE & \textbf{57.5} & \textbf{43.5} & \textbf{71.0} & \textbf{60.0}  & \textbf{88.3} & \textbf{69.4} &\textbf{70.7} & \textbf{48.9} & 41.8 & \textbf{20.6}   \\
LDE with CUHK03 &- & - & - & -  & 88.0 &70.7 &72.0 & 50.9 & 43.4 & 21.8   \\
\hline
\end{tabular}
\end{table*}

\begin{figure*}[t]
\begin{center}
\includegraphics[width=0.8\linewidth,height=0.25\linewidth]{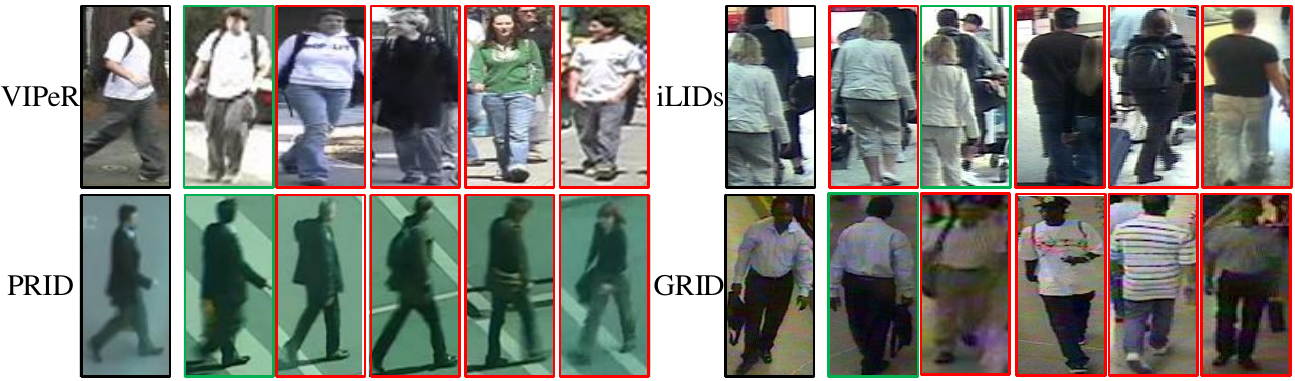}
\end{center}
   \caption{Top-5 ranking visualization results of our LDE on four small-scale target domains. The black, green and red boxes indicate the queries, correct and wrong results, respectively}
\label{visualres}
\end{figure*}

\textbf{3) Performance on the Collected OWD.} We compare the proposed LDE with some powerful DG ReID methods under three different settings in Table \ref{perform_owd} on the collected OWD. Under the first multi-source DG setting, the models are evaluated on the testing set of OWD following open-scene protocol, and the training data is the combination of the training set of Market1501, CUHK03, and MSMT17. We could observe that LDE achieves competitive performance compared with DG ReID methods. For example, the mAP performance is improved by 5\% compared with powerful MetaBIN that simulates unsuccessful normalization scenarios by meta-learning to achieve strong generalization ability. Under the second traditional person ReID setting, we exploit the training set of OWD to train DG models and then verify the performance on the testing data of OWD. Compared with these powerful DG methods, the proposed LDE could achieve better performance. These results illustrate that these advanced DG ReID methods may ignore some discriminative information when focusing on the domain-invariant features. LDE could not only extract domain-invariant features but also have the potential to retain useful domain-specific information. Under the third single-domain DG setting, we train the models on the training set of the collected OWD dataset following open-scene protocol and then generalize them to different large-scale target domains, i.e., Market1501, Dukemtmc, and MSMT17, respectively. On the Market1501 dataset, the mAP accuracy increases by 4.3\% compared with powerful CBN that exploits camera information to alleviate the distribution gap. Through simple but effective feature-level domain expansion, LDE could promote domain-invariant learning and capture discriminative features under different settings. Compared with the powerful DG ReID methods, LDE could provide a more comprehensive generalization ability. Further, on the setting denoted by `3' in Table \ref{perform_owd}, we provide extra multi-source DG experiments, i.e., combining the developed OWD benchmark and the existing CUHK03 dataset as the source domain. It can be seen that training the LDE with OWD and other public data combined could yield better results. For example, the Rank-1 accuracy is improved by 1.6\% on large-scale MSMT17.


\subsection{Visualization Results}

\textbf{1) Attention Maps.} In order to have an in-depth insight on domain decoupling in LDE, we visualize the heatmaps of the learned identy-relevant features and domain-relevant features. Specifically,
we adopt Grad-cam \cite{selvaraju2017grad} on CUHK03 dataset to visualize the attention maps in Fig.~\ref{attention}. 
It is clear that the identity-relevant features focus on the pedestrian torso part, while the domain-relevant features focus on background regions with ambiguous pixels removed. This brings two merits: on one hand, the identity-relevant features are focused, which thus improves accuracy. On the other hand, the domain-relevant features are purified by removing ambiguous pixels, providing accurate domain style for domain expansion, which thus improves generalization. 

\textbf{2) Ranking Results.} Fig. \ref{visualres} shows the Top 5 ranking results of our LDE on four small-scale target benchmarks. Note that there is only one correct matching sample in the gallery set for each query image. We observe that although the query and gallery images are from different viewpoints and show different appearances, LDE recalls ground-truth images in top 5 rankings. In particular, our method returns a wrong Rank-1 result on iLIDs. This is rational because two persons are overlapped in the query image and the model focuses on the foreground while overlooking the background. Therefore, the foreground person is ranked as top 1, which, instead, reflects the objective of LDE.  

\begin{figure}[t]
\begin{center}
\includegraphics[width=0.8\linewidth,height=0.3\linewidth]{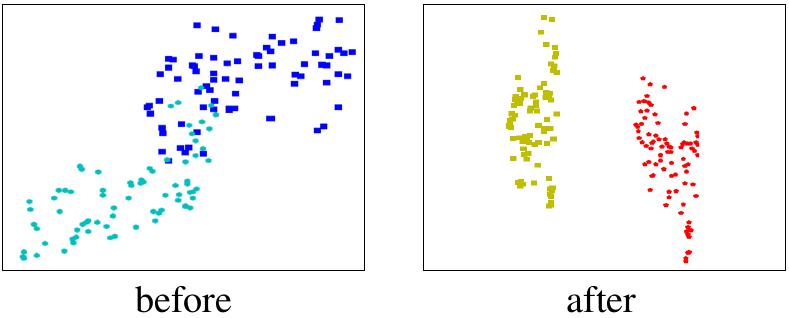}
\end{center}
   \caption{The t-SNE visualization before and after domain expansion. Cyan and gold scatters represent identity-relevant features, and blue and red scatters represent domain-relevant features}
\label{tsne}
\end{figure}

\textbf{3) Feature Distribution.} In order to show the effect of domain decoupling (step 1) and domain expansion (step 2), we exploit t-SNE~\cite{van2008visualizing} to visualize features on iLIDs after step 1 and step 2 in Fig.~\ref{tsne}, respectively. We observe from Fig.~\ref{tsne}(left) that the domain-relevant features can be decoupled from the identity-relevant features, which show the effect of step 1. After domain expansion in step 2, they become more separable and 
the ability of the model to extract domain-invariant features is further enhanced.

\subsection{Ablation Study of LDE}

\begin{table}[t]
\centering
\caption{Ablation study. The best results are bolded}\label{Ablation}
\setlength{\tabcolsep}{0.7mm}
\begin{tabular}{lcccc}
\toprule%
& \multicolumn{4}{@{}@{}c@{}}{Market1501} \\\cmidrule{2-5}%
Methods & Rank-1 & Rank-5  & Rank-10 & mAP \\
\midrule
baseline & 67.97 & 82.46 & 86.80 & 40.49\\
+ Step 1 without MSLS & 75.25 & 87.48 & 91.61 & 49.37\\
+ Step 1 with MSLS & 76.48 & 87.50 & 90.94 & 50.34\\
+ Step 2 & 68.21 & 82.26 & 87.15 & 41.75\\
LDE (+Step 1+Step 2) & \textbf{78.38} & \textbf{89.28} & \textbf{92.28} & \textbf{52.75} \\
\hline
\end{tabular}
\end{table}

\begin{figure}[t]
\begin{center}
\includegraphics[width=0.9\linewidth,height=0.4\linewidth]{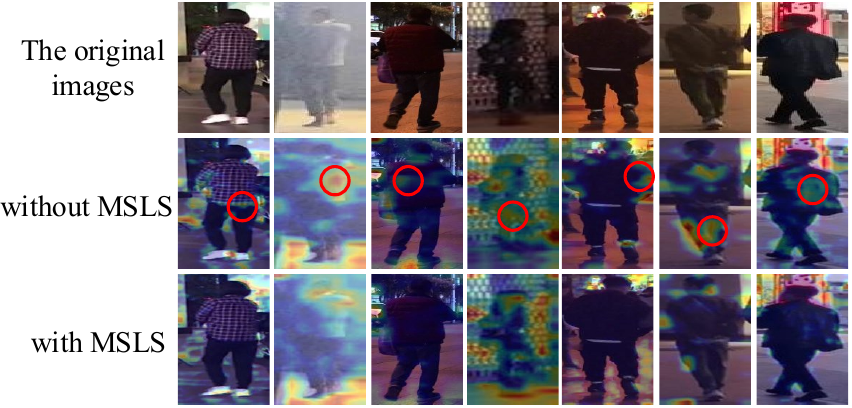}
\end{center}
   \caption{The comparison of domain-relevant feature heatmaps of night samples with and without the MSLS, where red circles indicate the identity information suppressed by the MSLS module}
\label{figure13}
\end{figure}

\begin{figure}[t]
\begin{center}
\includegraphics[width=0.9\linewidth,height=0.5\linewidth]{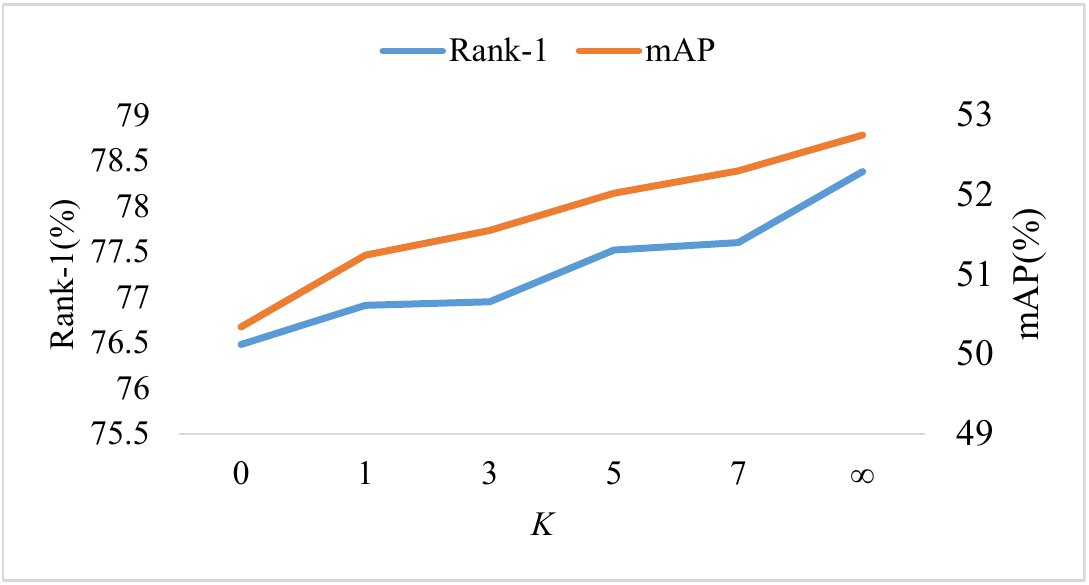}
\end{center}
   \caption{Ablation study about the values of $K$}
\label{figure14}
\end{figure}
We conduct ablation experiments on Market1501 to study the influence of each component in LDE. Specifically, the source domain is the training data of Dukemtmc+Cuhk03+MSMT17, and the target domain is Market1501. The results are presented in Table \ref{Ablation}. We observe that after deploying multiple DDMs (i.e., step 1 without MSLS), the Rank-1 accuracy and mAP are improved by 7.28\% and 8.88\%, respectively. These results indicate the necessity of decoupling domain-relevant and identity-relevant information from original features. After deploying MSLS (i.e., step 1 with MSLS), the Rank-1 performance on the Market1501 dataset is improved by 1.23\% and mAP improved by 0.97\%. MSLS could further improve performance, which proves that discriminative identity-relevant features could be mined effectively by the lifting-suppressing scheme. The baseline achieves limited 1.26\% mAP gains by only deploying Step 2 individually, this shows that expanding features towards meaningless directions does not bring too much contribution. Therefore, capturing meaningful transformation directions is of great importance to achieve domain expansion effectively. Finally, with two steps jointly trained, they complement each other and promote domain-invariant learning. These results verify the effectiveness of each component in the LDE.

In Fig.~\ref{figure13}, we visualize the domain-relevant feature heatmaps of hard night samples with and without the MSLS module. The identity information (red circles in the second row) in domain-relevant features could be suppressed by the MSLS. Therefore, MSLS can serve as a supplement to DDMs to provide pure domain-relevant features for more effective Domain Expansion in Step 2. Besides, compared to the daytime samples in Fig. 10, the retrieval of nighttime samples is relatively difficult, so the domain-relevant features captured by LDE are relatively limited.

Fig.~\ref{figure14} illustrates the performance comparison when the number of augmentation $K$ is varied. When the value of $K$ is small, the accuracy gain is limited. As the $K$ increases, the Rank-1 and mAP performance has a gradual improvement. When $K$ tends to be infinite (i.e., the proposed LDE), it achieves the best state in DG performance. This demonstrates the effectiveness of the proposed Domain Expansion for implicit augmentation at the feature level.


\section{Conclusion and Outlook}

This paper studies challenging person ReID in dynamic wild applications by developing an open-world benchmark and an efficient DG model simultaneously. Our major contribution is the large-scale benchmark dataset i.e., OWD. Compared with previous ReID benchmarks, this new dataset has several distinct features, such as diverse open-world collection scenes, diverse lighting variations, diverse person status, etc. OWD shows inspiring potential for promoting the community of person ReID, pedestrian detection and tracking. The developed OWD shows more challenges approaching realistic scenes, which may thus arouse more progress in generalized methodology of person ReID. 
We believe that under the open-world and cross-spatial-temporal setting proposed by OWD, a number of unresolved issues and untapped space can be amended and explored. In addition, we present a DG person ReID method i.e., LDE, which conducts domain expansion and promotes domain invariant learning, so as to improve both accuracy and generalization. Comprehensive evaluations based on a number of experiments show that OWD and LDE have superior transferable effects on close-scene and open-scene protocols. 
The contributions of this work facilitate the generalization ability of deep ReID models and make a crucial step forward towards practical applications.

Although this work makes critical efforts and progress, it is far from the grand goal toward open-world and dynamic wild applications. In the case of clothing changes, OWD provides relatively limited improvement. However, in practical surveillance system, the open-world and cross-spatial-temporal settings are usually accompanied by clothing changes of a target person, a practical but intractable challenge of person ReID. Therefore, it is crucial to enrich clothing changes of the same person so as to train a clothing-insensitive model. However, it is impractical to manually collect cloth-changing real-world data due to person privacy sensitivity, but with the developed OWD, some deep learning based clothes-changing technology can be leveraged for semi-synthesis in future work. 
We share the OWD benchmark and LDE codes with the community toward having much greater progress in practical applications and further narrowing the gap between vision technology and dynamic wild scenes. 
Moreover, a large-scale benchmark full of cloth-changing data containing open-world and semi-synthetic data can be a suitable solution to bridge this gap. 

%

\textbf{Data Availability}

An open-world dataset, i.e., OWD, is developed, which is available from the corresponding author upon acceptance and reasonable request. This paper also uses other 11 public datasets for person re-identification (\cite{6909421}, \cite{zheng2015scalable}, \cite{zheng2017unlabeled}, \cite{8578114}, \cite{shu2021large}, \cite{Barbosa2018Looking}, \cite{bak2018domain}, \cite{sun2019dissecting}, \cite{wang2020surpassing}, \cite{zhang2021unrealperson} and \cite{wang2022cloning}).


%

\appendices


%
%

\ifCLASSOPTIONcaptionsoff
  \newpage
\fi



%

\bibliographystyle{IEEEtrans}
\bibliography{sn-bibliography}

\begin{thebibliography}{10}
\providecommand{\url}[1]{#1}
\csname url@samestyle\endcsname
\providecommand{\newblock}{\relax}
\providecommand{\bibinfo}[2]{#2}
\providecommand{\BIBentrySTDinterwordspacing}{\spaceskip=0pt\relax}
\providecommand{\BIBentryALTinterwordstretchfactor}{4}
\providecommand{\BIBentryALTinterwordspacing}{\spaceskip=\fontdimen2\font plus
\BIBentryALTinterwordstretchfactor\fontdimen3\font minus
  \fontdimen4\font\relax}
\providecommand{\BIBforeignlanguage}[2]{{%
\expandafter\ifx\csname l@#1\endcsname\relax
\typeout{** WARNING: IEEEtranS.bst: No hyphenation pattern has been}%
\typeout{** loaded for the language `#1'. Using the pattern for}%
\typeout{** the default language instead.}%
\else
\language=\csname l@#1\endcsname
\fi
#2}}
\providecommand{\BIBdecl}{\relax}
\BIBdecl

\bibitem{2021Person30K}
Y.~Bai, J.~Jiao, W.~Ce, J.~Liu, Y.~Lou, X.~Feng, and L.-Y. Duan, ``Person30k: A
  dual-meta generalization network for person re-identification,'' in
  \emph{CVPR}, 2021, pp. 2123--2132.

\bibitem{bak2018domain}
S.~Bak, P.~Carr, and J.-F. Lalonde, ``Domain adaptation through synthesis for
  unsupervised person re-identification,'' in \emph{ECCV}, 2018, pp. 189--205.

\bibitem{baltieri20113dpes}
D.~Baltieri, R.~Vezzani, and R.~Cucchiara, ``3dpes: 3d people dataset for
  surveillance and forensics,'' in \emph{ACM workshop on Human gesture and
  behavior understanding}, 2011, pp. 59--64.

\bibitem{Barbosa2018Looking}
I.~B. Barbosa, M.~Cristani, B.~Caputo, A.~Rognhaugen, and T.~Theoharis,
  ``Looking beyond appearances: Synthetic training data for deep cnns in
  re-identification,'' \emph{Computer Vision and Image Understanding}, vol.
  167, pp. 50--62, 2018.

\bibitem{Blanchard2011GeneralizingFS}
G.~Blanchard, G.~Lee, and C.~Scott, ``Generalizing from several related
  classification tasks to a new unlabeled sample,'' in \emph{NeurIPS}, vol.~24,
  2011, pp. 2178--2186.

\bibitem{chen2021dual}
P.~Chen, P.~Dai, J.~Liu, F.~Zheng, M.~Xu, Q.~Tian, and R.~Ji, ``Dual
  distribution alignment network for generalizable person re-identification,''
  in \emph{AAAI}, vol.~35, no.~2, 2021, pp. 1054--1062.

\bibitem{2020Salience}
X.~Chen, C.~Fu, Y.~Zhao, F.~Zheng, J.~Song, R.~Ji, and Y.~Yang,
  ``Salience-guided cascaded suppression network for person
  re-identification,'' in \emph{CVPR}, 2020, pp. 3300--3310.

\bibitem{cheng2011custom}
D.~S. Cheng, M.~Cristani, M.~Stoppa, L.~Bazzani, and V.~Murino, ``Custom
  pictorial structures for re-identification.'' in \emph{BMVC}, vol.~1,
  no.~2.\hskip 1em plus 0.5em minus 0.4em\relax Citeseer, 2011, p.~6.

\bibitem{choi2021meta}
S.~Choi, T.~Kim, M.~Jeong, H.~Park, and C.~Kim, ``Meta batch-instance
  normalization for generalizable person re-identification,'' in \emph{CVPR},
  2021, pp. 3425--3435.

\bibitem{dai2021generalizable}
Y.~Dai, X.~Li, J.~Liu, Z.~Tong, and L.-Y. Duan, ``Generalizable person
  re-identification with relevance-aware mixture of experts,'' in \emph{CVPR},
  2021, pp. 16\,145--16\,154.

\bibitem{fu2020unsupervised}
D.~Fu, D.~Chen, J.~Bao, H.~Yang, L.~Yuan, L.~Zhang, H.~Li, and D.~Chen,
  ``Unsupervised pre-training for person re-identification,'' in \emph{CVPR},
  2021, pp. 14\,750--14\,759.

\bibitem{2008Viewpoint}
D.~Gray and H.~Tao, ``Viewpoint invariant pedestrian recognition with an
  ensemble of localized features,'' in \emph{ECCV}, 2008.

\bibitem{7780459}
K.~{He}, X.~{Zhang}, S.~{Ren}, and J.~{Sun}, ``Deep residual learning for image
  recognition,'' in \emph{CVPR}, 2016, pp. 770--778.

\bibitem{he2020fastreid}
L.~He, X.~Liao, W.~Liu, X.~Liu, P.~Cheng, and T.~Mei, ``Fastreid: a pytorch
  toolbox for real-world person re-identification,'' \emph{arXiv preprint
  arXiv:2006.02631}, vol.~1, no.~6, 2020.

\bibitem{hermans2017defense}
A.~Hermans, L.~Beyer, and B.~Leibe, ``In defense of the triplet loss for person
  re-identification,'' \emph{arXiv preprint arXiv:1703.07737}, 2017.

\bibitem{Hirzer2011PersonRB}
M.~Hirzer, C.~Beleznai, P.~M. Roth, and H.~Bischof, ``Person re-identification
  by descriptive and discriminative classification,'' in \emph{Scandinavian
  conference on Image analysis}, 2011, pp. 91--102.

\bibitem{8578843}
J.~Hu, L.~Shen, and G.~Sun, ``Squeeze-and-excitation networks,'' in
  \emph{CVPR}, 2018, pp. 7132--7141.

\bibitem{huang2018unifying}
Q.~Huang, Y.~Xiong, and D.~Lin, ``Unifying identification and context learning
  for person recognition,'' in \emph{CVPR}, 2018, pp. 2217--2225.

\bibitem{huang2019celebrities}
Y.~Huang, Q.~Wu, J.~Xu, and Y.~Zhong, ``Celebrities-reid: A benchmark for
  clothes variation in long-term person re-identification,'' in
  \emph{IJCNN}.\hskip 1em plus 0.5em minus 0.4em\relax IEEE, 2019, pp. 1--8.

\bibitem{huang2019beyond}
Y.~Huang, J.~Xu, Q.~Wu, Y.~Zhong, P.~Zhang, and Z.~Zhang, ``Beyond scalar
  neuron: Adopting vector-neuron capsules for long-term person
  re-identification,'' \emph{IEEE TCSVT}, vol.~30, no.~10, pp. 3459--3471,
  2019.

\bibitem{huang2022learning}
Y.~Huang, X.~Fu, L.~Li, and Z.-J. Zha, ``Learning degradation-invariant
  representation for robust real-world person re-identification,''
  \emph{International Journal of Computer Vision}, vol. 130, no.~11, pp.
  2770--2796, 2022.

\bibitem{jia2019frustratingly}
J.~Jia, Q.~Ruan, and T.~M. Hospedales, ``Frustratingly easy person
  re-identification: Generalizing person re-id in practice,'' \emph{arXiv
  preprint arXiv:1905.03422}, 2019.

\bibitem{9157711}
X.~Jin, C.~Lan, W.~Zeng, Z.~Chen, and L.~Zhang, ``Style normalization and
  restitution for generalizable person re-identification,'' in \emph{CVPR},
  2020, pp. 3140--3149.

\bibitem{jin2020style}
------, ``Style normalization and restitution for generalizable person
  re-identification,'' in \emph{CVPR}, 2020, pp. 3143--3152.

\bibitem{li2017learning}
D.~Li, Y.~Yang, Y.-Z. Song, and T.~M. Hospedales, ``Learning to generalize:
  Meta-learning for domain generalization,'' in \emph{AAAI}, vol.~32, no.~1,
  2018.

\bibitem{li2021simple}
P.~Li, D.~Li, W.~Li, S.~Gong, Y.~Fu, and T.~M. Hospedales, ``A simple feature
  augmentation for domain generalization,'' in \emph{ICCV}, 2021, pp.
  8886--8895.

\bibitem{6909421}
W.~{Li}, R.~{Zhao}, T.~{Xiao}, and X.~{Wang}, ``Deepreid: Deep filter pairing
  neural network for person re-identification,'' in \emph{CVPR}, 2014, pp.
  152--159.

\bibitem{6619305}
W.~Li and X.~Wang, ``Locally aligned feature transforms across views,'' in
  \emph{CVPR}, 2013, pp. 3594--3601.

\bibitem{2012Human}
W.~Li, R.~Zhao, and X.~Wang, ``Human reidentification with transferred metric
  learning,'' in \emph{ACCV}, 2012, pp. 31--44.

\bibitem{li2020scalable}
W.~Li, X.~Zhu, and S.~Gong, ``Scalable person re-identification by harmonious
  attention,'' \emph{International Journal of Computer Vision}, vol. 128,
  no.~6, pp. 1635--1653, 2020.

\bibitem{Li2023StyleControllableGP}
Y.~Li, J.~Song, H.~Ni, and H.~T. Shen, ``Style-controllable generalized person
  re-identification,'' in \emph{ACMMM}, 2023, pp. 7912--7921.

\bibitem{liao2020interpretable}
S.~Liao and L.~Shao, ``Interpretable and generalizable person re-identification
  with query-adaptive convolution and temporal lifting,'' in \emph{ECCV}.\hskip
  1em plus 0.5em minus 0.4em\relax Springer, 2020, pp. 456--474.

\bibitem{liao2022graph}
------, ``Graph sampling based deep metric learning for generalizable person
  re-identification,'' in \emph{CVPR}, 2022, pp. 7359--7368.

\bibitem{2014Person}
C.~C. Loy, C.~Liu, and S.~Gong, ``Person re-identification by manifold
  ranking,'' in \emph{ICIP}, 2013, pp. 3567--3571.

\bibitem{Luo_2019_CVPR_Workshops}
H.~Luo, Y.~Gu, X.~Liao, S.~Lai, and W.~Jiang, ``Bag of tricks and a strong
  baseline for deep person re-identification,'' in \emph{CVPRW}, 2019, pp.
  0--0.

\bibitem{luo2019switchable}
P.~Luo, R.~Zhang, J.~Ren, Z.~Peng, and J.~Li, ``Switchable normalization for
  learning-to-normalize deep representation,'' \emph{IEEE TPAMI}, vol.~43,
  no.~2, pp. 712--728, 2019.

\bibitem{ma2016orientation}
L.~Ma, H.~Liu, L.~Hu, C.~Wang, and Q.~Sun, ``Orientation driven bag of
  appearances for person re-identification,'' \emph{arXiv preprint
  arXiv:1605.02464}, 2016.

\bibitem{martinel2012re}
N.~Martinel and C.~Micheloni, ``Re-identify people in wide area camera
  network,'' in \emph{CVPRW}.\hskip 1em plus 0.5em minus 0.4em\relax IEEE,
  2012, pp. 31--36.

\bibitem{9880010}
H.~Ni, J.~Song, X.~Luo, F.~Zheng, W.~Li, and H.~T. Shen, ``Meta distribution
  alignment for generalizable person re-identification,'' in \emph{CVPR}, 2022,
  pp. 2477--2486.

\bibitem{nichol2018firstorder}
A.~Nichol, J.~Achiam, and J.~Schulman, ``On first-order meta-learning
  algorithms,'' \emph{arXiv preprint arXiv:1803.02999}, 2018.

\bibitem{pan2018two}
X.~Pan, P.~Luo, J.~Shi, and X.~Tang, ``Two at once: Enhancing learning and
  generalization capacities via ibn-net,'' in \emph{ECCV}, 2018.

\bibitem{qian2020long}
X.~Qian, W.~Wang, L.~Zhang, F.~Zhu, Y.~Fu, T.~Xiang, Y.-G. Jiang, and X.~Xue,
  ``Long-term cloth-changing person re-identification,'' in \emph{ACCV}, 2020.

\bibitem{Act2Param}
S.~Qiao, C.~Liu, W.~Shen, and A.~L. Yuille, ``Few-shot image recognition by
  predicting parameters from activations,'' in \emph{CVPR}, 2018, pp.
  7229--7238.

\bibitem{rao2019learning}
Y.~Rao, J.~Lu, and J.~Zhou, ``Learning discriminative aggregation network for
  video-based face recognition and person re-identification,''
  \emph{International Journal of Computer Vision}, vol. 127, pp. 701--718,
  2019.

\bibitem{russakovsky2015imagenet}
O.~Russakovsky, J.~Deng, H.~Su, J.~Krause, S.~Satheesh, S.~Ma, Z.~Huang,
  A.~Karpathy, A.~Khosla, M.~Bernstein \emph{et~al.}, ``Imagenet large scale
  visual recognition challenge,'' \emph{International Journal of Computer
  Vision}, vol. 115, no.~3, pp. 211--252, 2015.

\bibitem{selvaraju2017grad}
R.~R. Selvaraju, M.~Cogswell, A.~Das, R.~Vedantam, D.~Parikh, and D.~Batra,
  ``Grad-cam: Visual explanations from deep networks via gradient-based
  localization,'' in \emph{ICCV}, 2017, pp. 618--626.

\bibitem{shankar2018generalizing}
S.~Shankar, V.~Piratla, S.~Chakrabarti, S.~Chaudhuri, P.~Jyothi, and
  S.~Sarawagi, ``Generalizing across domains via cross-gradient training,''
  \emph{arXiv preprint arXiv:1804.10745}, 2018.

\bibitem{shu2021large}
X.~Shu, X.~Wang, X.~Zang, S.~Zhang, Y.~Chen, G.~Li, and Q.~Tian, ``Large-scale
  spatio-temporal person re-identification: Algorithms and benchmark,''
  \emph{IEEE TCSVT}, vol.~32, no.~7, pp. 4390--4403, 2021.

\bibitem{shu2021open}
Y.~Shu, Z.~Cao, C.~Wang, J.~Wang, and M.~Long, ``Open domain generalization
  with domain-augmented meta-learning,'' in \emph{CVPR}, 2021, pp. 9624--9633.

\bibitem{2019Generalizable}
J.~Song, Y.~Yang, Y.-Z. Song, T.~Xiang, and T.~M. Hospedales, ``Generalizable
  person re-identification by domain-invariant mapping network,'' in
  \emph{CVPR}, 2019, pp. 719--728.

\bibitem{sun2019dissecting}
X.~Sun and L.~Zheng, ``Dissecting person re-identification from the viewpoint
  of viewpoint,'' in \emph{CVPR}, 2019, pp. 608--617.

\bibitem{sun2018models}
Y.~Sun, L.~Zheng, Y.~Yang, Q.~Tian, and S.~Wang, ``Beyond part models: Person
  retrieval with refined part pooling (and a strong convolutional baseline),''
  in \emph{ECCV}, 2018, pp. 480--496.

\bibitem{tamura2019augmented}
M.~Tamura and T.~Murakami, ``Augmented hard example mining for generalizable
  person re-identification,'' \emph{arXiv preprint arXiv:1910.05280}, 2019.

\bibitem{van2008visualizing}
L.~Van~der Maaten and G.~Hinton, ``Visualizing data using t-sne.''
  \emph{Journal of Machine Learning Research}, vol.~9, no.~11, 2008.

\bibitem{wang2020weakly}
G.~Wang, G.~Wang, X.~Zhang, J.~Lai, Z.~Yu, and L.~Lin, ``Weakly supervised
  person re-id: Differentiable graphical learning and a new benchmark,''
  \emph{IEEE TNNLS}, vol.~32, no.~5, pp. 2142--2156, 2020.

\bibitem{2018Learning}
G.~Wang, Y.~Yuan, X.~Chen, J.~Li, and X.~Zhou, ``Learning discriminative
  features with multiple granularities for person re-identification,'' in
  \emph{ACMMM}, 2018, pp. 274--282.

\bibitem{wang2022cloning}
Y.~Wang, X.~Liang, and S.~Liao, ``Cloning outfits from real-world images to 3d
  characters for generalizable person re-identification,'' in \emph{CVPR},
  2022, pp. 4900--4909.

\bibitem{wang2020surpassing}
Y.~Wang, S.~Liao, and L.~Shao, ``Surpassing real-world source training data:
  Random 3d characters for generalizable person re-identification,'' in
  \emph{ACMMM}, 2020, pp. 3422--3430.

\bibitem{wang2019implicit}
Y.~Wang, X.~Pan, S.~Song, H.~Zhang, G.~Huang, and C.~Wu, ``Implicit semantic
  data augmentation for deep networks,'' in \emph{NeurIPS}, vol.~32, 2019.

\bibitem{8578114}
L.~Wei, S.~Zhang, W.~Gao, and Q.~Tian, ``Person transfer gan to bridge domain
  gap for person re-identification,'' in \emph{CVPR}, 2018, pp. 79--88.

\bibitem{2018CBAM}
S.~Woo, J.~Park, J.~Y. Lee, and I.~S. Kweon, ``Cbam: Convolutional block
  attention module,'' in \emph{ECCV}, 2018, pp. 3--19.

\bibitem{xiao2017joint}
T.~Xiao, S.~Li, B.~Wang, L.~Lin, and X.~Wang, ``Joint detection and
  identification feature learning for person search,'' in \emph{CVPR}, 2017,
  pp. 3415--3424.

\bibitem{xu2023deepchange}
P.~Xu and X.~Zhu, ``Deepchange: A long-term person re-identification benchmark
  with clothes change,'' in \emph{ICCV}, 2023, pp. 11\,196--11\,205.

\bibitem{yang2019person}
Q.~Yang, A.~Wu, and W.-S. Zheng, ``Person re-identification by contour sketch
  under moderate clothing change,'' \emph{IEEE TPAMI}, vol.~43, no.~6, pp.
  2029--2046, 2019.

\bibitem{arxiv20reidsurvey}
M.~Ye, J.~Shen, G.~Lin, T.~Xiang, L.~Shao, and S.~C.~H. Hoi, ``Deep learning
  for person re-identification: A survey and outlook,'' \emph{arXiv preprint
  arXiv:2001.04193}, 2020.

\bibitem{yin2020fine}
J.~Yin, A.~Wu, and W.-S. Zheng, ``Fine-grained person re-identification,''
  \emph{International Journal of Computer Vision}, vol. 128, pp. 1654--1672,
  2020.

\bibitem{zhai2023population}
Y.~Zhai, P.~Peng, M.~Jia, S.~Li, W.~Chen, X.~Gao, and Y.~Tian,
  ``Population-based evolutionary gaming for unsupervised person
  re-identification,'' \emph{International Journal of Computer Vision}, vol.
  131, no.~1, pp. 1--25, 2023.

\bibitem{2019Night}
J.~Zhang, Y.~Yuan, and Q.~Wang, ``Night person re-identification and a
  benchmark,'' \emph{IEEE Access}, vol.~PP, no.~99, pp. 1--1, 2019.

\bibitem{zhang2020does}
L.~Zhang, Z.~Deng, K.~Kawaguchi, A.~Ghorbani, and J.~Zou, ``How does mixup help
  with robustness and generalization?'' \emph{arXiv preprint arXiv:2010.04819},
  2020.

\bibitem{zhang2015beyond}
N.~Zhang, M.~Paluri, Y.~Taigman, R.~Fergus, and L.~Bourdev, ``Beyond frontal
  faces: Improving person recognition using multiple cues,'' in \emph{CVPR},
  2015, pp. 4804--4813.

\bibitem{zhang2021unrealperson}
T.~Zhang, L.~Xie, L.~Wei, Z.~Zhuang, Y.~Zhang, B.~Li, and Q.~Tian,
  ``Unrealperson: An adaptive pipeline towards costless person
  re-identification,'' in \emph{CVPR}, 2021, pp. 11\,506--11\,515.

\bibitem{zhang2017alignedreid}
X.~Zhang, H.~Luo, X.~Fan, W.~Xiang, Y.~Sun, Q.~Xiao, W.~Jiang, C.~Zhang, and
  J.~Sun, ``Alignedreid: Surpassing human-level performance in person
  re-identification,'' \emph{arXiv preprint arXiv:1711.08184}, 2017.

\bibitem{zhang2020relation}
Z.~Zhang, C.~Lan, W.~Zeng, X.~Jin, and Z.~Chen, ``Relation-aware global
  attention for person re-identification,'' in \emph{CVPR}, 2020, pp.
  3186--3195.

\bibitem{zhao2022revisiting}
J.~Zhao, Y.~Zhao, X.~Chen, and J.~Li, ``Revisiting stochastic learning for
  generalizable person re-identification,'' in \emph{ACMMM}, 2022, pp.
  1758--1768.

\bibitem{zhao2021learning}
Y.~Zhao, Z.~Zhong, F.~Yang, Z.~Luo, Y.~Lin, S.~Li, and S.~Nicu, ``Learning to
  generalize unseen domains via memory-based multi-source meta-learning for
  person re-identification,'' in \emph{CVPR}, 2021, pp. 6277--6286.

\bibitem{zheng2015scalable}
L.~Zheng, L.~Shen, L.~Tian, S.~Wang, J.~Wang, and Q.~Tian, ``Scalable person
  re-identification: A benchmark,'' in \emph{ICCV}, 2015, pp. 1116--1124.

\bibitem{zheng2017person}
L.~Zheng, H.~Zhang, S.~Sun, M.~Chandraker, Y.~Yang, and Q.~Tian, ``Person
  re-identification in the wild,'' in \emph{CVPR}, 2017, pp. 1367--1376.

\bibitem{zheng2018rpifield}
M.~Zheng, S.~Karanam, and R.~J. Radke, ``Rpifield: A new dataset for temporally
  evaluating person re-identification,'' in \emph{CVPRW}, 2018, pp. 1893--1895.

\bibitem{2009Associating}
W.-S. Zheng, S.~Gong, and T.~Xiang, ``Associating groups of people.'' in
  \emph{BMVC}, vol.~2, no.~6, 2009, pp. 1--11.

\bibitem{zheng2017unlabeled}
Z.~Zheng, L.~Zheng, and Y.~Yang, ``Unlabeled samples generated by gan improve
  the person re-identification baseline in vitro,'' in \emph{ICCV}, 2017, pp.
  3754--3762.

\bibitem{zhou2021learning}
K.~Zhou, Y.~Yang, A.~Cavallaro, and T.~Xiang, ``Learning generalisable
  omni-scale representations for person re-identification,'' \emph{IEEE TPAMI},
  vol.~44, no.~9, pp. 5056--5069, 2021.

\bibitem{zhou2021domain}
K.~Zhou, Y.~Yang, Y.~Qiao, and T.~Xiang, ``Domain generalization with
  mixstyle,'' \emph{arXiv preprint arXiv:2104.02008}, 2021.

\bibitem{zhu2019intra}
X.~Zhu, X.~Zhu, M.~Li, V.~Murino, and S.~Gong, ``Intra-camera supervised person
  re-identification: A new benchmark,'' in \emph{ICCVW}, 2019, pp. 0--0.

\bibitem{zhuang2020rethinking}
Z.~Zhuang, L.~Wei, L.~Xie, T.~Zhang, H.~Zhang, H.~Wu, H.~Ai, and Q.~Tian,
  ``Rethinking the distribution gap of person re-identification with
  camera-based batch normalization,'' in \emph{ECCV}.\hskip 1em plus 0.5em
  minus 0.4em\relax Springer, 2020, pp. 140--157.

\end{thebibliography}
\end{document}